\pdfoutput=1
\documentclass[sigconf]{acmart}
\usepackage{xcolor,colortbl}
\usepackage{booktabs}
\usepackage{multirow}
\usepackage{rotating}
\usepackage{diagbox}
\usepackage{array}
\usepackage{enumerate}
\usepackage{amsfonts}
\usepackage{amsmath}
\usepackage{bookmark}
\usepackage{enumitem}
\usepackage{bbding}
\usepackage{cleveref}
\usepackage{arydshln}

\AtBeginDocument{%
  }

\definecolor{new_olive}{HTML}{8b9656}
\newcommand{\correct}[0]{\small\color{new_olive}\CheckmarkBold}
\definecolor{new_pink}{HTML}{e096a7}
\newcommand{\wrong}[0]{\small\color{new_pink}\XSolidBrush}

\definecolor{orangered}{HTML}{fa585b}
\newcommand{\red}[1]{\textbf{\color{orangered}{#1}}}

\usepackage{balance}

\copyrightyear{2025}
\acmYear{2025}
\setcopyright{acmlicensed}\acmConference[MM '25]{Proceedings of the 33rd ACM International Conference on Multimedia}{October 27--31, 2025}{Dublin, Ireland}
\acmBooktitle{Proceedings of the 33rd ACM International Conference on Multimedia (MM '25), October 27--31, 2025, Dublin, Ireland}
\acmDOI{10.1145/3746027.3754862}
\acmISBN{979-8-4007-2035-2/2025/10}
\acmSubmissionID{859}

\begin{document}
\title{DetectAnyLLM: Towards Generalizable and Robust Detection of Machine-Generated Text Across Domains and Models}
\begin{abstract}
The rapid advancement of large language models (LLMs) has drawn urgent attention to the task of machine-generated text detection (MGTD).
However, existing approaches struggle in complex real-world scenarios: zero-shot detectors rely heavily on scoring model's output distribution while training-based detectors are often constrained by overfitting to the training data, limiting generalization.
We found that the performance bottleneck of training-based detectors stems from the misalignment between training objective and task needs.
To address this, we propose \textbf{Direct Discrepancy Learning (DDL)}, a novel optimization strategy that directly optimizes the detector with task-oriented knowledge.
DDL enables the detector to better capture the core semantics of the detection task, thereby enhancing both robustness and generalization.
Built upon this, we introduce \textbf{DetectAnyLLM}, a unified detection framework that achieves state-of-the-art MGTD performance across diverse LLMs.
To ensure a reliable evaluation, we construct \textbf{MIRAGE}, the most diverse multi-task MGTD benchmark.
MIRAGE samples human-written texts from 10 corpora across 5 text-domains, which are then re-generated or revised using 17 cutting-edge LLMs, covering a wide spectrum of proprietary models and textual styles.
Extensive experiments on MIRAGE reveal the limitations of existing methods in complex environment.
In contrast, DetectAnyLLM consistently outperforms them, achieving over a 70\% performance improvement under the same training data and base scoring model, underscoring the effectiveness of our DDL.
Project page: ~\url{https://fjc2005.github.io/detectanyllm}.
\end{abstract}

\author{Jiachen Fu}
\affiliation{%
  \institution{VCIP, CS, Nankai University}
  \city{Tianjin}
  \country{China}
}
\email{fujiachen2005@gmail.com}

\author{Chun-Le Guo}
\authornote{Corresponding Author.}
\affiliation{%
  \institution{VCIP, CS, Nankai University}
  \city{Tianjin}
  \country{China}
}
\affiliation{%
  \institution{NKIARI}
  \city{Shenzhen Futian}
  \country{China}
}
\email{guochunle@nankai.edu.cn}

\author{Chongyi Li}
\authornote{Project Lead.}
\affiliation{%
  \institution{VCIP, CS, Nankai University}
  \city{Tianjin}
  \country{China}
}
\affiliation{%
  \institution{NKIARI}
  \city{Shenzhen Futian}
  \country{China}
}
\email{lichongyi@nankai.edu.cn}

\renewcommand{\shortauthors}{Jiachen Fu, Chun-Le Guo, \& Chongyi Li}



\begin{teaserfigure}
  \centering
  \includegraphics[width=\textwidth]{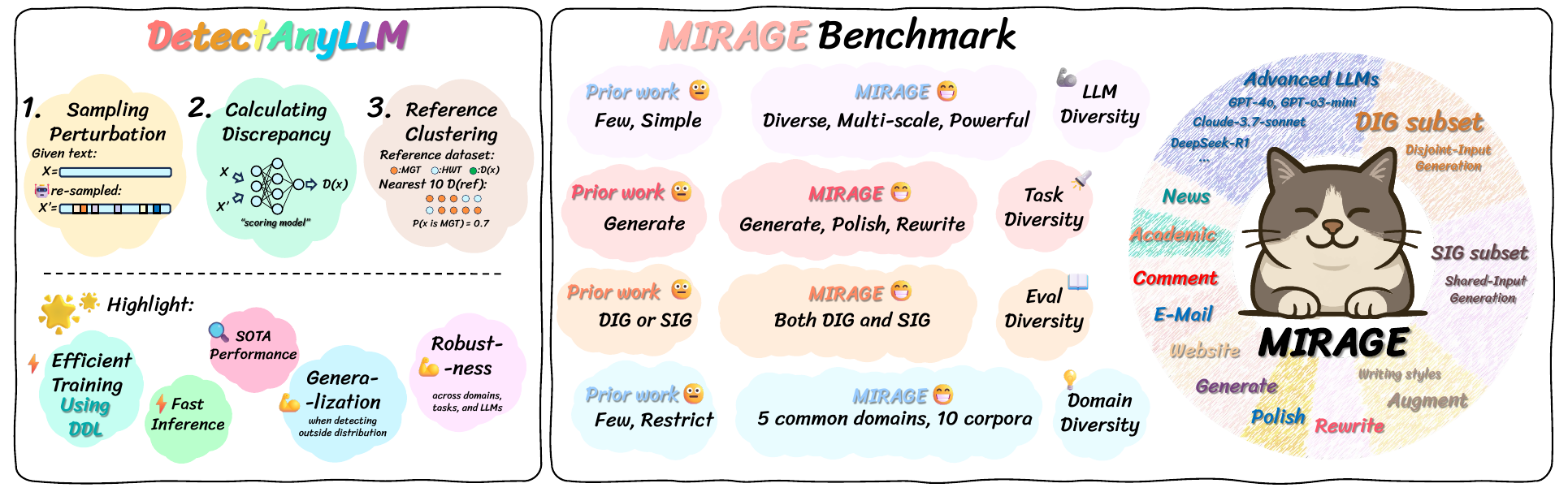}
  \caption{Left: Our DetectAnyLLM achieves high efficiency, strong robustness, and impressive generalization through a three-step process: \textit{sampling perturbation}, \textit{calculating discrepancy}, and \textit{reference clustering}. Right: Our MIRAGE benchmark emphasizes diversity across domains, tasks, evaluation scenarios, and source LLMs, enabling comprehensive and robust evaluation.}
  \Description{Overview of our work. }
  \label{fig:teaser}
\end{teaserfigure}

\maketitle

\section{Introduction}
Advanced \textit{Large Language Models (LLMs)}~\cite{gpt4, gpt4o, deepseekr1, deepseekv3, gpto1, gemini} can easily generate text nearly indistinguishable from human writing~\cite{indistinguish_1, indistinguish_2}.
If misused, it could pose serious risks to society~\cite{fakenews}.
In response to such concern, the task of \textit{Machine-Generated Text Detection (MGTD)} has emerged~\cite{logrank, entropy, likelihood, LLM-DetectAIve, resitrict_embeds, survey_automatic}.
MGTD is a binary classification task designed to distinguish whether a given text is written by humans or generated (or revised) by a machine.

In this study, we consider detection of both \textit{Machine-Generated Text (MGT)} and \textit{Machine-Revised Text (MRT)}, where MRT refers to text that polished or rewritten by a model based on \textit{Human-Written Text (HWT)}.
Our study focuses on black-box detection, which better reflects real-world application than white-box settings.

Several MGTD methods have been proposed~\cite{logrank, likelihood, entropy, detectgpt, fastdetectgpt, text_fluoroscopy, binoculars, instrinsic} based on the assumption that the token probability distributions are distinct between MGT and HWT.
Most of them leverage pre-trained language models, referred to as \textit{scoring models}, to estimate the token probabilities of a given text and compute classification metrics for distinguishing.

Existing MGTD methods can be categorized into \textit{zero-shot methods}~\cite{entropy, lrrandnpr, detectgpt, fastdetectgpt, hart} and \textit{training-based methods}~\cite{logrank, imbd}.

Zero-shot methods typically rely on the inherent capabilities of scoring models~\cite{glimpse}. However, these models are often relatively small, with limited knowledge and simple output patterns. Thus, when detecting texts deviate from their inherent distribution, such methods often struggle to achieve reliable performance.

Training-based methods use \textit{supervised fine-tuned (SFT)}~\cite{sft, logrank} or \textit{preference learning}~\cite{orpo, dpo, imbd} to align the scoring model's output distribution with that of models who built the training data.

While such approach improves detection performance for that specific models, it seems difficult to generalize this detection knowledge to models outside the training data~\cite{fastdetectgpt, survey_necessity_methods_futuredirect, survey_science}.

We point out that both SFT and preference learning steer the scoring model toward mimicking the generators rather than being optimized directly for detection.

\textit{In other words, the training goal of previous training-based methods is model-oriented, rather than task-oriented.}

This leads to the fact that the scoring model can only learn the knowledge of the generators of the training data, but cannot learn the knowledge of the detection task directly.
Ultimately, it jeopardizes the generalization and robustness of the detector.

We propose \textbf{Direct Discrepancy Learning (DDL)}, a novel optimization strategy that enables the model to \textbf{learn to be a detector rather than another language model} by directly optimizing the scoring model with the output classification metric.
DDL originates from the idea of freeing the scoring model from its identity as a language model and designs a task-oriented loss function so that the scoring model can directly learn the intrinsic knowledge of MGTD instead of simply fitting the distribution of training data.

Furthermore, as the fusion of prior approaches~\cite{fastdetectgpt, imbd} and DDL, we propose \textbf{DetectAnyLLM}, a unified MGTD framework.
DetectAnyLLM achieves efficient and robust detection through three steps comprising \textit{re-sampling}, \textit{discrepancy calculation}, and \textit{reference clustering}.
Such a framework distills the core insights of existing methods~\cite{detectgpt, fastdetectgpt} while leveraging DDL to enhance the model’s generalization capabilities and improve detection robustness.

\begin{table*}[htbp]
    \centering
    \caption{Comparison between MIRAGE and existing MGTD benchmark datasets. ``Size" is the capacity of the test set. ``SIG" denotes Shared-Input Generation and ``DIG" denotes Disjoint-Input Generation. ``Commercial" refers to the use of frontier proprietary LLMs (e.g., GPT-4o). MIRAGE is the most diverse benchmark in terms of domain, tasks, and source LLMs. MIRAGE leverages powerful proprietary LLMs to generate and revise text, increasing the difficulty of detection and the realism of evaluation, enabling a more faithful evaluation of detector robustness. Furthermore, MIRAGE introduces a novel dual-scenario evaluation strategy —DIG and SIG— allowing more comprehensive assessment of both accuracy and generalization capacity.}
    \renewcommand{\arraystretch}{1.25}
    \resizebox{\linewidth}{!}{
    \begin{tabular}{l|ccc|c|ccc|ccc}
    \hline

    \hline

    \hline
    &\multicolumn{3}{c|}{\textbf{Data Statistic}}&\multicolumn{1}{c|}{\textbf{LLMs}}&\multicolumn{3}{c|}{\textbf{MGT Tasks}}&\multicolumn{3}{c}{\textbf{Other}}\\
    \textbf{Benchmark}& Size & Domain Coverage & Corpus & Commercial & Generate & Polish & Rewrite & Aug. & SIG & DIG\\
    \hline
    TuringBench~\cite{turingbench}& 40K & News & 3 & \wrong & \correct & \wrong & \wrong & \wrong & \wrong & \correct\\
    HC3~\cite{hc3}& 85K & QA/Comment/Academic & 5 & 1  & \correct & \wrong & \wrong & \wrong & - & - \\
    M4~\cite{m4bench}& 24.5K & QA/Comment/Academic/News & 11 & 2 & \correct & \wrong & \wrong & \correct & \correct & \wrong\\
    MAGE~\cite{mage}& 29K & QA/Comment/News/Academic/Story & 10 & 3 & \correct & \wrong & \wrong & \correct & \correct & \wrong \\
    RAID~\cite{raid}& 628.7K & News/Academic/Comment/Literature& 11& 3 & \correct & \wrong & \correct & \correct & \correct & \wrong\\
    DetectRL~\cite{detectrl}&134.4K& Academic/Comment & 4 & 2&\correct&\wrong&\correct&\correct&\correct&\wrong \\
    HART~\cite{hart}&16K& News/Literature/Academic& 4 & 4 & \correct & \correct &\correct &\correct & \wrong & \correct \\    
    \rowcolor[HTML]{fff5f4}
    \textbf{MIRAGE (ours)}&93.8K &Academic/Comment/Email/News/Website & 10 & 13 & \correct & \correct & \correct & \correct & \correct & \correct \\
    \hline

    \hline

    \hline
    \end{tabular}
    }
    \label{tab:benchmarks}
\end{table*}

\begin{figure}[t]
    \centering
    \includegraphics[width=\linewidth]{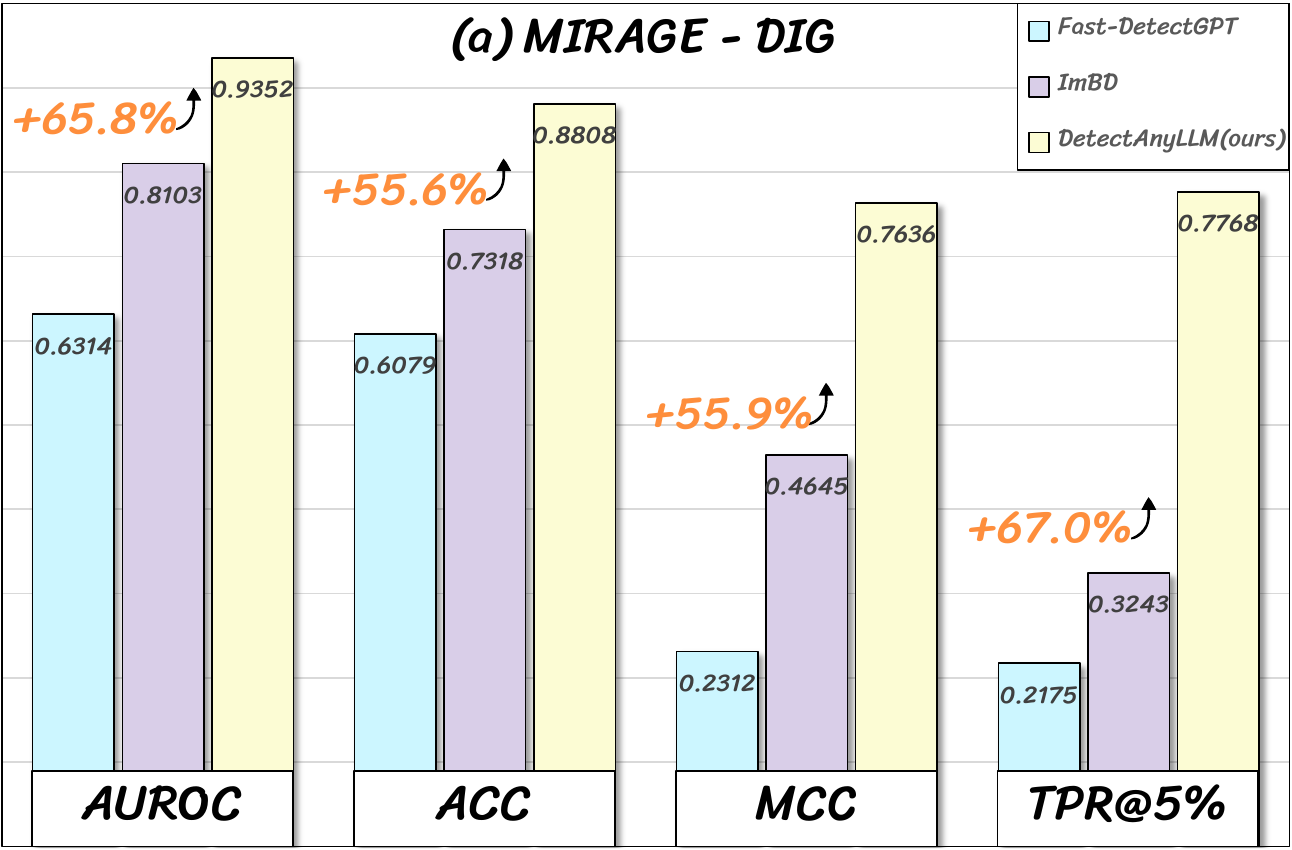}
    \caption{Performance on MIRAGE-DIG for DetectAnyLLM and State-Of-The-Art methods, including Fast-DetectGPT~\cite{fastdetectgpt}, and ImBD~\cite{imbd}. Imp.: $(new-old)/(1.0-old)$.}
    \Description{Performance Comparison on MIRAGE. The DIG subset reflects abilities across domains, and the SIG subset reflects abilities across generators.}
    \label{fig:performance_comparison}
\end{figure}

Despite various MGTD studies have emerged, there remains a lack of comprehensive benchmarks~\cite{survey_necessity_methods_futuredirect}.
Existing benchmarks~\cite{mage, mgtbench, detectrl, roft} suffer from several significant deficiencies: \textbf{1) Limited focus on MRT:} Most benchmark datasets, such as MGTBench~\cite{mgtbench}, focus solely on MGT while neglecting the detection of MRT. \textbf{2) Narrow range of source LLMs:} Most benchmarks rely on small-scale, open-source models, whereas real-world applications often involve advanced, proprietary LLMs such as GPT-4o~\cite{gpt4o} and Claude~\cite{claude}. \textbf{3) Restricted domain coverage:} Benchmarks such as HC3~\cite{hc3} sample text from only one or a few domains, neglecting the domain sensitivity of machine-generated text.
These deficiencies highlight a significant gap between evaluation and real-world applications.
Although some recent studies~\cite{survey_necessity_methods_futuredirect, hart} have recognized such problems, their datasets remain insufficiently comprehensive.

To facilitate a comprehensive evaluation, we construct \textbf{MIRAGE}, the largest multi-task MGTD benchmark that provides the richest variety of proprietary LLMs and the most comprehensive text domains in MGTD research.
As shown in ~\Cref{tab:benchmarks}, MIRAGE samples text from \textbf{10 corpora} in \textbf{5 common domains}, and uses \textbf{17 advanced mainstream LLMs} for text generation or revision, creating over \textbf{93K HWT-MGT pairs}.
MIRAGE establishes a more realistic and reliable evaluation standard, bridging the gap between research and real-world applications.

Though existing detection methods have demonstrated seemingly outstanding performance (AUC > 0.9) on previous benchmarks~\cite{imbd, detectrl}, they exhibit significant weaknesses when evaluated on MIRAGE, as ~\Cref{fig:performance_comparison} shown.
This reveals the generalization and robustness of prior methods require substantial improvement. 
In contrast, DetectAnyLLM still performs well, achieving an average of \red{0.9352} AUROC and \red{0.7636} MCC on the MIRAGE-DIG subset.
Such performance powerfully demonstrates the efficiency and superior generalization of DDL.

Our contributions can be summarized into the following points:

\begin{itemize}[leftmargin=*]
    \item We propose \textbf{Direct Discrepancy Learning (DDL)}, a novel task-oriented optimization method that improves generalization and robustness with fewer resources and no extra data.
    \item We construct \textbf{MIRAGE}, a comprehensive MGTD benchmark covering diverse domains, tasks, and novelly focuses on using proprietary LLMs, resulting a more realistic evaluation.
    \item We present \textbf{DetectAnyLLM}, unifying prior works and DDL, achieving up to \textbf{70\%} performance gains and realizing a generalizable and robust detection across domains and models.
\end{itemize}

\section{Related Work}
\subsection{Zero-shot Detector}
Previous MGTD research emphasized zero-shot detection due to concerns about overfitting during training~\cite{outdomain, overfit}. 
Early methods like GLTR~\cite{entropy} leveraged text entropy to detect machine content, while others used likelihood- or ranking-based approaches~\cite{likelihood, logrank}.
Recently, DetectGPT~\cite{detectgpt} provides a novel view for MGTD, it distinguishes MGT from HWT using \textit{perturbation}, \textit{scoring}, and \textit{probability curvature estimation}.
Fast-DetectGPT~\cite{fastdetectgpt} improves the perturbation step, significantly accelerating the detection process without sacrificing performance.
Despite progress, zero-shot methods remain constrained by their dependence on the scoring model’s output distribution, as shown by Glimpse~\cite{glimpse}, which demonstrated performance improvement through stronger scoring models.

\subsection{Training-based Detector}
The training-based detector fine-tunes the scoring model on specific training data.
An early representative work is RoBERTa-OpenAI-Detector~\cite{likelihood}, the researchers fine-tune RoBERTa~\cite{roberta} models using GPT-2~\cite{gpt2}-generated data, performing well on detecting GPT-generated text.
RADAR~\cite{radar} incorporates adversarial training~\cite{gan} to enhance MGTD robustness and uses PPO~\cite{ppo} to optimize the generator. 
More recently, ImBD~\cite{imbd} utilizes DPO~\cite{dpo} to optimize the scoring model based on the Fast-DetectGPT~\cite{fastdetectgpt} framework, aiming to help the scoring model better capture the style features of the training data.

Despite these advancements, most methods simply focus on training the scoring model to approximate the source model’s distribution rather than developing a dedicated MGTD detector. 
This introduces constraints to the scoring model during training, which are detrimental to the MGTD task.

\subsection{MGTD Benchmark}
Early benchmarks like Turingbench~\cite{turingbench} focused on news articles generated by neural models, while the emergence of ChatGPT~\cite{instructGPT} shifted attention to LLM-generated text, exemplified by MGTBench\cite{mgtbench} and HC3~\cite{hc3}. 
Later efforts such as MAGE~\cite{mage}, MULTITuDE~\cite{multitude}, and M4~\cite{m4bench} explored open-domain and multilingual detection.
RAID~\cite{raid} novelly introduced decoding strategy considerations to strengthen evaluation robustness, while DetectRL~\cite{detectrl} examined vulnerabilities from a writing-attack perspective.
However, most of these benchmarks rely on open-source models (indicating limited variousity) and focus mainly on MGT, overlooking more common real-world applications involving MRT, thus limits their applicability to real-world contexts.

HART~\cite{hart} marked progress by incorporating both MGT and MRT using six advanced LLMs (only four proprietary LLMs), but it remains limited in generator diversity and domain scope.
In this study, we scale up the number of generators to 17, where 13 are proprietary LLMs and 4 are advanced open-source LLMs, covering nearly all major LLMs used in real-world applications.
Moreover, we sample HWT~\cite{xlsum, bigpatent} from five distinct domains and generate both MGT and MRT, ensuring a more comprehensive and representative evaluation.
To advance MGTD research and enable fairer comparisons, we advocate for the adoption of a unified benchmark to ensure consistency in evaluation standards. We hope MIRAGE will serve as a valuable step toward achieving this goal.
\section{DetectAnyLLM Framework}
DetectAnyLLM builds upon Fast-DetectGPT~\cite{fastdetectgpt}, which determines whether a text is MGT by measuring the log-probability discrepancy between the original text and its perturbed variants~\cite{detectgpt}.
This method involves three key steps: \textit{1) re-sampling the given text}, \textit{2) computing the discrepancy between original text and re-sampled text}, and \textit{3) making a decision using the discrepancy}.
DDL is utilized to train the scoring model to enhance steps 1) and 2) so that the detector can more easily distinguish between MGT and HWT.
In \Cref{3.1}, we describe how the log-probability discrepancy is calculated.
Next, in \Cref{3.2}, we explain the motivation behind our improvements to this detection process, along with the specific designs we introduce.
Finally, in \Cref{3.3}, we detail how discrepancy is ultimately used for MGTD within our proposed framework.

\subsection{Preliminary}\label{3.1}

\noindent \textbf{Basic Hypothesis. }
Machine-generated text tends to consist of high-probability tokens at each position, whereas human-written text has greater variability.
Although sampling strategies like top-k and top-p introduce some randomness, LLMs still generally select tokens with relatively high probabilities.
Thus, features in the probability distribution of tokens can serve as useful cues for distinguishing machine-generated text from human-written.

\noindent \textbf{Probability Discrepancy. }
Given a text $x$ and a scoring model $f_\theta$, when using a language model $q_\phi$ to produce perturbations, the \textit{probability discrepancy} (i.e., probability curvature)~\cite{detectgpt} can be expressed as:
\begin{equation}
    d(x, f_\theta, q_\phi) = \log f_\theta(x) - \mathbb E_{\tilde{x} \sim q_\phi(\cdot|x)}[\log f_\theta(\tilde{x})],
    \label{3.1.1}
\end{equation}
where $\tilde{x}$ is the perturbed version of $x$ by $q_\phi$.

Based on the hypothesis, machine-generated text $x_m$ tends to have a high log-probability, whereas its perturbed version $\tilde{x_m}$ shows a lower log-probability.
In contrast, human-written text $x_h$ generally has a lower log-probability.
When perturbed, $\tilde{x_h}$ tends to show an increase in log-probability, as the perturbation process replaces words in $x_h$ with higher-likelihood alternatives according to the model.
Thus, we expect to achieve:
\begin{equation}
    d(x_m, f_\theta, q_\phi) > d(x_h, f_\theta, q_\phi).
    \label{3.1.2}
\end{equation}
This inequality forms the basis of MGTD~\cite{detectgpt, fastdetectgpt, imbd}.

When calculating this discrepancy, achieving $f_\theta$ is straightforward, allowing $\log f_\theta(x)$ to be efficiently computed.
However, since the log-probabilities are computed using Markov-Chain, even a small perturbation requires recalculating the entire chain.
Thus, estimating the expectation of the log-probability of $\tilde{x}$ is complex.

\noindent \textbf{Conditional Probability. }~\cite{fastdetectgpt} is a biased yet computationally efficient estimation of the original probability:
\begin{equation}
    f_\theta(\tilde{x}) = \prod_{i}f_\theta(\tilde{x}_i|\tilde{x}_{<i}) \sim \prod_{i}f_\theta(\tilde{x}_i|x_{<i}) = f_\theta(\tilde{x}|x).
    \label{3.1.3}
\end{equation}
By introducing Eq.~\eqref{3.1.3}, the probability discrepancy in Eq.~\eqref{3.1.1} can be further reformulated to the \textit{conditional probability discrepancy}:
\begin{equation}
    d_c(x, f_\theta, q_\phi) = \frac{\log f_\theta(x|x) - \tilde{\mu}}{\tilde{\sigma}},
    \label{3.1.4}
\end{equation}
where
\begin{equation}
\begin{aligned}
    \tilde{\mu} &= \mathbb E_{\tilde{x}\sim q_\phi(\tilde{x}|x)}[\log f_\theta (\tilde{x}|x)],\\
    \tilde{\sigma}^2 &= \mathbb E_{\tilde{x}\sim q_\phi(\tilde{x}|x)}[\log f_\theta (\tilde{x}|x) - \tilde{\mu}^2].
\end{aligned}
\label{3.1.5}
\end{equation}
Noticing that a normalization item $\tilde{\sigma}$ is added into the discrepancy function, we further explore how the $\tilde{\sigma}$ affects performance in ~\Cref{ablation_on_sigma}.
%

\noindent \textbf{Re-sample text.} 
Given a sentence consisting of $s$ tokens, we use the model $q_\phi$ to compute $q_\phi(t|x_{<i})$ for $i$ from $1$ to $s$, where $t$ represents for token.
This results in a tensor $lprobs$ of shape $(s, v)$, where $v$ denotes the vocabulary size of $q_\phi$.
With such a tensor, we can efficiently generate $n$ re-sampled samples with only a single line of PyTorch code.

For the original version of probability discrepancy~\cite{detectgpt}, both perturbation generation and discrepancy estimation require calculating a whole Markov-Chain for $n$ times.
This leads to a time complexity of $O(n\times s)$.

By introducing conditional probability~\cite{fastdetectgpt}, the resampling approach can replace the perturbation step.
Under this formulation, both generating $n$ samples and computing the discrepancy require running the Markov-Chain only \textbf{once}.
As a result, the time complexity is reduced to $O(s)$.

\begin{figure}[t]
    \centering
    \includegraphics[width=0.65\linewidth]{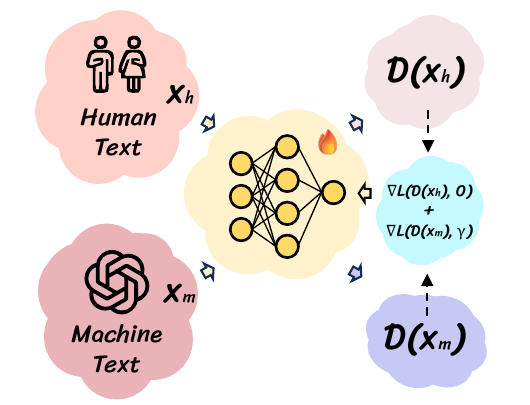}
    \caption{Overview of Direct Discrepancy Learning. The scoring model receives paired HWT-MGT data and computes the discrepancy for each. We optimize it to minimize the HWT's discrepancy while maximizing the MGT's.}
    \Description{Overview of Direct Discrepancy Learning. }
    \label{fig:ddl}
\end{figure}

\subsection{Optimizing by Direct Discrepancy Learning}\label{3.2}
As shown in Eq.~\eqref{3.1.4} and Eq.~\eqref{3.1.2}, the key to enhance the detector's performance is to increase the distribution difference of the conditional probability discrepancy between MGT and HWT estimated by the scoring model.

While ImBD~\cite{imbd} has achieved significant performance gains by incorporating \textit{Direct Preference Optimization (DPO)}~\cite{dpo} to optimize the scoring model, we argue that DPO is not the optimal optimization method for the MGTD task.

\noindent \textbf{DPO.}
~\cite{dpo} is derived from the optimization objective of \textit{Proximal Policy Optimization (PPO)}~\cite{ppo}, which is:
\begin{equation}
    \mathop{max}_\theta \mathbb{E}_{x\sim f_\theta(x)}[r(x)] - \beta\mathbb{D}_{KL}[f_\theta(x) \mid \mid f_{ref}(x)],
    \label{3.2.1}
\end{equation}
where $x$ is a text sampled from the scoring model $f_\theta$'s distribution, and $r$ is a reward function that can judge whether this sample is bad or good.
By analyzing and re-parameterizing this optimization objective, we can obtain DPO's optimization objective:
\begin{equation}
    \mathop{max}_\theta \mathbb{E}_{x_m, x_h \sim \mathop{D}}[\log \sigma(\beta\log\frac{f_\theta(x_m)}{f_{ref}(x_m)}-\beta\log\frac{f_\theta(x_h)}{f_{ref}(x_h)})],
    \label{3.2.6}
\end{equation}
where $x_m$ denotes MGT and $x_h$ stands for HWT.
$f_{ref}$ is a reference model, usually the original $f_\theta$.
The detailed derivation process will be presented in the supplementary material.

\noindent \textbf{Motivated by redundant KL-regularization. }\label{redundant KL-regularization}
The KL term between $f_\theta$ and $f_{ref}$ is explicitly added to the optimization objective in PPO~\cite{ppo}, and its weight is adjusted via $\beta$, as shown in Eq. \eqref{3.2.1}.
While in DPO~\cite{dpo}, as Eq. 
\eqref{3.2.6} shown, such regularization is implicitly embedded in the optimization objective, and its strength can also be adjusted by $\beta$.
ImBD~\cite{imbd} directly adopts Eq.\eqref{3.2.6} as its loss function and leverages paired MGT-HWT data to optimize the scoring model $f_\theta$.
The KL-regularization forces the scoring model to retain its internal knowledge while learning preferences.

\textit{This leads us to question: for the MGTD task, what is the significance of retaining the original knowledge of the scoring model during training?}

Since we have introduced training, our direct objective should be enable the scoring model to better capture the knowledge of the MGTD task.
Fundamentally, we hope that the training process will teach the scoring model how to become a detector.
However, the KL-regularization drastically shifts this objective: from learning the intrinsic knowledge of the MGTD task to aligning the scoring model with the distribution of training data.
This shifts the training process from learn a detector to mimic a language model, thereby misleading the scoring model.

\noindent \textbf{Direct Discrepancy Learning. }
Based on the reasoning above, we remove the KL-regularization in the optimization objective.
Thus, the optimization goal can be re-written as:
\begin{equation}
    \mathop{max}_{\theta} \mathbb{E}_{x\sim \mathop{D}}[r(x)].
    \label{3.3.1}
\end{equation}
We further design a simple but task-oriented reward objective $r(x)$, defined as:
\begin{equation}
    r(x) = \begin{cases}
    -\Vert \gamma - d_c(x, f_\theta, q_\phi) \Vert_1, & \text{when $x$ is $x_m$,}\\
    -\Vert d_c(x, f_\theta, q_\phi) \Vert_1, & \text{when $x$ is $x_h$.}
    \end{cases}
    \label{3.3.2}
\end{equation}
where $\gamma$ is an hyper-parameter.
This reward function is designed based on the conclusions discussed in Section~\ref{3.1}, that is the discrepancy of human-written text $x_h$ tends to be low (close to 0) while the discrepancy of machine-generated text $x_m$ tends to be positive.
The parameter $\gamma$ is introduced to control how positive the discrepancy of $x_m$ should be.
In our experiment, $\gamma$ is arbitrarily chosen.
As shown in ~\Cref{tab:ablation_gamma}, an experiment on the impact of the value of $\gamma$ shows that the model's performance is not particularly sensitive to this choice, indicating a level of robustness to variations in $\gamma$. 
In practice, our input consists of paired HWT-MGT data. We set $q_\phi = f_\theta$ following the ImBD~\cite{imbd}'s setting, which allows us to use the scoring model’s output for optimization: 
\begin{equation}
    \mathop{min}_\theta \mathbb{E}_{x_m, x_h \sim \mathop{D}}(\Vert d_c(x_h, f_\theta, f_\theta)\Vert_1 + \Vert \gamma - d_c(x_m, f_\theta, f_\theta)\Vert_1).
    \label{3.3.3}
\end{equation}
We call this optimization method as \textit{Direct Discrepancy Learning (DDL)}, as it helps the scoring model directly learn the expected conditional probability discrepancy of both MGT and HWT.

By removing the KL-regularization, the scoring model can essentially forget its identity as a language model.
Furthermore, the reward function based on the discrepancy $d_c$, which incorporates a task-oriented prior, can help the scoring model to directly learn the inherent knowledge of MGTD.
Specifically, $d_c$ for HWT approaches $0$, while $d_c$ for MGT is positive.

\subsection{Detecting by Reference Clustering}\label{3.3}
We use \textit{Reference Clustering} to achieve the transition from $d_c(x)$ to $p_{m}(x)$.
Specifically, this algorithm is designed to estimate the probability of a given value belonging to a specific distribution, consisting of: \textit{data aggregation} and \textit{probability estimation}.

\noindent \textbf{Data Aggregation. }
We first collect a certain number of MGT texts as the MGT reference dataset $M$, and an approximately equal number of HWT texts as the HWT reference dataset $H$.
Then, we employ the scoring model $f_\theta$, which will be used for detection, to respectively compute the conditional probability discrepancy $d_c$ for each text in $M$ and $H$.
Thereby , we can obtain the conditional probability discrepancy distribution $D_m$ and $D_h$ of the texts in $M$ and $H$ under scored by $f_\theta$.

\noindent \textbf{Probability Estimation. }
We select the value in $M\cup H$ that is $k_{th}$ closest to the target value $d_c(x)$ as the search window $\delta$:
\begin{equation}
    \delta = sorted(\{\Vert d_c(x_{ref}) - d_c(x)\Vert_1 \mid x_{ref}\in M \cup H\})[k],
    \label{3.3.4}
\end{equation}
where $k$ is a hyper-parameter that should be determined by the size of reference dataset.
For a larger reference dataset, a larger $k$ is better as it can provide higher precision of $p_{m}(x)$.

Then, we count the number of MGT texts and HWT texts within the window range:
\begin{equation}
    \begin{aligned}
        cnt_m = \sum_{d\in D_m}I(d_c(x) - \delta < d < d_c(x) + \delta),\\
        cnt_h = \sum_{d\in D_h}I(d_c(x) - \delta < d < d_c(x) + \delta).
    \end{aligned}
    \label{3.3.5}
\end{equation}
Finally, we estimate the probability that text $x$ belongs to MGT using the local statistical ratio:
\begin{equation}
    p_m(x) = \frac{cnt_m}{cnt_m + cnt_h}.
    \label{3.3.6}
\end{equation}

Since the window $\delta$ is adaptively determined by the data distribution, this method can maintain stability under different data densities, thereby improving the robustness of real-world MGTD.

\begin{figure*}
    \centering
    \includegraphics[width=\linewidth]{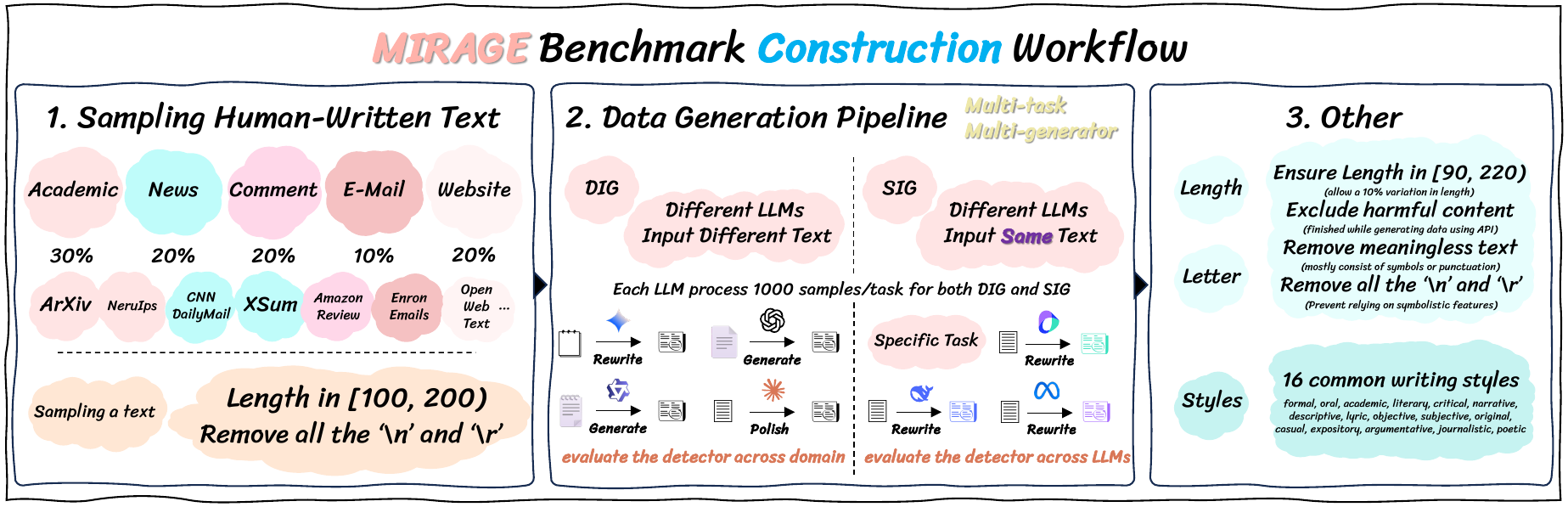}
    \caption{The MIRAGE benchmark construction workflow. MIRAGE consists of 93K HWT-MGT pairs, significantly demonstrating diversity of text-domain, source LLM, and generation task, while using writing style control as augmentation. }
    \Description{Workflow of MIRAGE Benchmark.}
    \label{fig:workflow}
\end{figure*}
\section{Proposed MIRAGE Benchmark}
Current benchmarks exhibit notable limitations in diversity of text domains~\cite{turingbench, detectrl}, coverage of source LLMs~\cite{m4bench, raid}, and evaluation tasks~\cite{mage, hc3}.

To facilitate a generalized evaluation that better reflects real-world application, we present the \textit{\textbf{M}ulti-domain \textbf{I}nclusive \textbf{R}ealistic \textbf{A}ssessment for machine \textbf{G}enerated text d\textbf{E}tection (\textbf{MIRAGE})} benchmark.
MIRAGE constitutes the most comprehensive multi-task MGTD evaluation framework to date, incorporating both generative and revisionary text across diverse domains, employing most advanced LLMs, including 13 proprietary and 4 open-source LLMs.

\subsection{Benchmark Construction}
 
\noindent \textbf{Multi-domain Sampling. }
Considering that LLMs exhibit varying performance across different text domains, MIRAGE samples HWT of 5 domains from 10 corpora.
Detail information is presented in Supplementary Material.

\noindent \textbf{Pre-Cleaning. }
We remove all the `\textbackslash n' character to prevent the detector from identifying MGT based on the presence of the `\textbackslash n' symbol. 
Subsequently, we filter out texts containing 100-200 words from these datasets to control for length-based detection biases.

\noindent \textbf{Inclusive MGT Tasks. }
Following established methodologies in the ~\cite{fastdetectgpt} and ~\cite{imbd}, we designed three distinct MGT tasks: \textit{Generate}, \textit{Polish}, and \textit{Rewrite}.
The \textit{Generate} task involves creating new text based on the first 30 tokens of an HWT.
The \textit{Polish} task refines an existing HWT while preserving its original details and meaning.
The \textit{Rewrite} task paraphrases a given HWT without altering its meaning or essential details.
The detailed prompt of each task will be presented in Supplementary Material.

\noindent \textbf{Realistic LLM Usage. }
In real-world applications, people typically rely on powerful proprietary LLMs to generate or revise text.
However, most existing benchmarks~\cite{turingbench, hc3, m4bench, mage, raid, detectrl} rely on open-source LLMs to build data, resulting in a gap between current evaluation and real-world applications.
To address this, MIRAGE incorporates 13 mainstream proprietary LLMs, as detailed in Supplementary Material.

Concurrently, recognizing the increasing deployment of high-performance open-source models in localized applications, we incorporated four advanced open-source LLMs~\cite{qwen2.5, llama3}, ensuring comprehensive coverage of the contemporary LLM ecosystem.

\noindent \textbf{Composition. }
We consider two distinct evaluation scenarios to better reflect real-world applications:

\textit{\textbf{Disjoint-Input Generation(DIG):}} Each LLM generates MGT or MRT based on a unique HWT. Detectors must distinguish between this machine output and its source HWT.

\textit{\textbf{Shared-Input Generation (SIG):}} Multiple LLMs generate MGT or MRT from the same HWT. Detectors must identify all machine outputs from a common input.

We design each LLM to generate 2,000 samples for each MGT task, equally distributed between DIG and SIG scenarios (1,000 samples each).
Both DIG and SIG follow the same domain distribution for consistency, as detailed in Supplementary Material.

Sampling begins with constructing domain level HWT datasets by proportionally merging source datasets within each domain.
Such dataset-mixing strategy eases dataset-bias by preventing oversampling from single dataset.

During implementation, SIG is treated as an independent ``model'' and incorporates alongside the 17 individual LLMs in the sampling process.
For each model (including SIG), we sequentially sample data from each domain dataset.
Once sampled, items are removed from corresponding domain datasets to maintain the distinction between DIG and SIG data.
Within each text domain, data is sampled continuously until the number of samples for that domain meets the requirements specified in Supplementary Material.
Once the data sampling for one text domain is complete, the process moves to the next, repeating until all text domains have been sampled.

This methodology produces the DIG dataset for each LLM and a comprehensive SIG dataset, which are subsequently combined to form the complete sample set for each LLM across all tasks.

\noindent \textbf{Data Augmentation. }
The language style is a key distinguishing feature between HWT and MGT, with a closer alignment to human language style posing a more challenging task for MGT detectors.
With this consideration, we introduce data augmentation in terms of LLM's language styles. 
Specifically, we incorporated the phrase ``in a <style> style" into the input prompt.
We manually select 16 different language styles, randomly choosing one during each LLM inference to achieve style-diversity.
This approach helps assess the robustness of detectors against language styles' attacks.

\noindent \textbf{Post Cleaning. }
After generating MGT or MRT from the above HWT data, we perform data cleaning on the generated data.
First, all the `\textbackslash n' and `\textbackslash r' are removed, to prevent detection from symbol's feature.
Next, we remove texts with fewer than 90 words or more than 220 words, to prevent the impact of text length variations on detection, and finally obtain the MIRAGE benchmark dataset.
The statistical results are presented in Supplementary Material.

\subsection{Evaluation Metrics}
Consistent with prior works~\cite{detectgpt, fastdetectgpt, imbd}, we adopt the \textit{Area Under the Receiver Operating Characteristic Curve (AUROC)} as the primary evaluation metric.
To assess the performance on specific threshold, we incorporate \textit{TPR at a 5\% false positive rate (TPR@5\%)} as a supplementary metric.
Furthermore, considering the MIRAGE-SIG is a class-imbalanced dataset, we additionally report the \textit{Matthews Correlation Coefficient (MCC)} and \textit{Balanced Accuracy} to provide a more comprehensive evaluation.
Together, this diverse set of metrics provides a comprehensive assessment of detector's performance, ensuring that the evaluation reflects both theoretical completeness and real-world applicability. 
\begin{table*}[htbp]
    \centering
    \caption{Results across three tasks (Generate, Polish, Rewrite) under two evaluation settings (MIRAGE-DIG and MIRAGE-SIG). All methods, except for RoBERTa-Base/Large, employ GPT-Neo-2.7B~\cite{gpt-neo} as the scoring model. Following the experiment settings of~\cite{imbd}, NPR~\cite{lrrandnpr} and DetectGPT~\cite{detectgpt} use T5-3B~\cite{t5} to generate perturbations, while Fast-DetectGPT~\cite{fastdetectgpt} utilizes GPT-J-6B~\cite{gpt-j} to generate samples. $\spadesuit$ indicates a training-based method, whereas $\diamondsuit$ denotes a method that requires multiple model invocations. "Imp." represents Improvement over previous SOTA, computed as $(new - old) / (1.0 - old)$. Metrics: AUROC, Balanced Accuracy, MCC, and TPR@5\%. DetectAnyLLM significantly outperforms all baselines across all tasks and settings.}
    \resizebox{\linewidth}{!}{
    \begin{tabular}{l|cccc|cccc|cccc}
    \hline

    \hline

    \hline
    \multicolumn{13}{c}{\textbf{MIRAGE-DIG (Disjoint-Input Generation)}}\\
    \hline

    \hline

    \hline
    \multirow{2}{*}{Methods}&\multicolumn{4}{c|}{Generate}&\multicolumn{4}{c|}{Polish}&\multicolumn{4}{c}{Rewrite} \\
    &  AUROC  &  Accuracy  &  MCC  &  TPR@5\%  &  AUROC  &  Accuracy  &  MCC  &  TPR@5\%  &  AUROC  &  Accuracy  &  MCC  &  TPR@5\%  \\
    \hline

    \hline
    Likelihood~\cite{likelihood} & 0.4936 & 0.5091 & 0.0183 & 0.0147 & 0.4653 & 0.5000 & 0.0000 & 0.0214 & 0.4337 & 0.5000 & 0.0000 & 0.0148 \\
    LogRank~\cite{logrank} & 0.4992 & 0.5128 & 0.0260 & 0.0220 & 0.4512 & 0.5000 & 0.0000 & 0.0195 & 0.4225 & 0.5000 & 0.0000 & 0.0132 \\
    Entropy~\cite{entropy} & 0.6522 & 0.6150 & 0.2543 & 0.1099 & 0.5543 & 0.5417 & 0.1247 & 0.0954 & 0.5805 & 0.5566 & 0.1650 & 0.1189 \\
    RoBERTa-Base~\cite{roberta} $\spadesuit$ & 0.5523 & 0.5397 & 0.1434 & 0.1250 & 0.4859 & 0.5010 & 0.0088 & 0.0460 & 0.5020 & 0.5049 & 0.0293 & 0.0569 \\
    RoBERTa-Large~\cite{roberta} $\spadesuit$ & 0.4716 & 0.5217 & 0.0842 & 0.0871 & 0.5171 & 0.5151 & 0.0340 & 0.0633 & 0.5570 & 0.5385 & 0.0864 & 0.0895 \\
    LRR~\cite{lrrandnpr} & 0.5215 & 0.5341 & 0.0777 & 0.0701 & 0.4081 & 0.5000 & 0.0000 & 0.0200 & 0.3930 & 0.5000 & 0.0000 & 0.0188 \\
    DNA-GPT~\cite{dna-gpt} $\diamondsuit$ & 0.5733 & 0.5595 & 0.1196 & 0.0776 & 0.4771 & 0.5004 & 0.0110 & 0.0309 & 0.4453 & 0.5001 & 0.0080 & 0.0251 \\
    NPR~\cite{lrrandnpr} $\diamondsuit$ & 0.6120 & 0.6140 & 0.2604 & 0.0191 & 0.5071 & 0.5370 & 0.1071 & 0.0318 & 0.4710 & 0.5201 & 0.0663 & 0.0226 \\
    DetectGPT~\cite{detectgpt} $\diamondsuit$ & 0.6402 & 0.6258 & 0.2758 & 0.0275 & 0.5469 & 0.5531 & 0.1328 & 0.0355 & 0.5061 & 0.5266 & 0.0826 & 0.0283 \\
    Fast-DetectGPT~\cite{fastdetectgpt} & 0.7768 & 0.7234 & 0.4628 & 0.4310 & 0.5720 & 0.5570 & 0.1293 & 0.1189 & 0.5455 & 0.5432 & 0.1015 & 0.1025 \\
    ImBD~\cite{imbd} $\spadesuit$ & 0.8597 & 0.7738 & 0.5497 & 0.4065 & 0.7888 & 0.7148 & 0.4300 & 0.2730 & 0.7825 & 0.7068 & 0.4139 & 0.2933 \\
    \hdashline
    
    \hdashline
    \rowcolor[HTML]{fff5f4}
    
    \textbf{DetectAnyLLM} (ours) $\spadesuit$ & \textbf{0.9525} & \textbf{0.8988} & \textbf{0.7975} & \textbf{0.7770} & \textbf{0.9297} & \textbf{0.8732} & \textbf{0.7487} & \textbf{0.7756} & \textbf{0.9234} & \textbf{0.8705} & \textbf{0.7447} & \textbf{0.7778} \\
    
    \rowcolor[HTML]{fff5f4}
    
    \textbf{Imp.} & \red{+66.14\%} & \red{+55.26\%} & \red{+55.03\%} & \red{+62.43\%} & \red{+66.71\%} & \red{+55.54\%} & \red{+55.91\%} & \red{+69.13\%} & \red{+64.78\%} & \red{+55.83\%} & \red{+56.44\%} & \red{+68.56\%} \\
    \hline

    \hline

    \hline
    \multicolumn{13}{c}{\textbf{MIRAGE-SIG (Shared-Input Generation)}}\\
    \hline

    \hline

    \hline
    \multirow{2}{*}{Methods}&\multicolumn{4}{c|}{Generate}&\multicolumn{4}{c|}{Polish}&\multicolumn{4}{c}{Rewrite} \\
    &  AUROC  &  Accuracy  &  MCC  &  TPR@5\%  &  AUROC  &  Accuracy  &  MCC  &  TPR@5\%  &  AUROC  &  Accuracy  &  MCC  &  TPR@5\%  \\
    \hline

    \hline
    Likelihood~\cite{likelihood} & 0.4968 & 0.5207 & 0.0196 & 0.0145 & 0.4599 & 0.5002 & 0.0030 & 0.0233 & 0.4319 & 0.5000 & 0.0000 & 0.0111 \\
    LogRank~\cite{logrank} & 0.5008 & 0.5183 & 0.0182 & 0.0186 & 0.4468 & 0.5000 & 0.0000 & 0.0211 & 0.4221 & 0.5000 & 0.0000 & 0.0118 \\  
    Entropy~\cite{entropy} & 0.6442 & 0.6123 & 0.1592 & 0.1074 & 0.5640 & 0.5439 & 0.0516 & 0.0946 & 0.5858 & 0.5645 & 0.0918 & 0.1198 \\  
    RoBERTa-Base~\cite{roberta} $\spadesuit$ & 0.5368 & 0.5392 & 0.0529 & 0.1101 & 0.4741 & 0.5011 & 0.0048 & 0.0395 & 0.5099 & 0.5122 & 0.0221 & 0.0668 \\
    RoBERTa-Large~\cite{roberta} $\spadesuit$ & 0.4703 & 0.5236 & 0.0417 & 0.0910 & 0.5150 & 0.5157 & 0.0283 & 0.0702 & 0.5576 & 0.5426 & 0.0405 & 0.0762 \\
    LRR~\cite{lrrandnpr} & 0.5214 & 0.5311 & 0.0314 & 0.0657 & 0.4076 & 0.5000 & 0.0000 & 0.0238 & 0.3978 & 0.5000 & 0.0000 & 0.0174 \\    
    DNA-GPT~\cite{dna-gpt} $\diamondsuit$ & 0.5759 & 0.5647 & 0.0603 & 0.0813 & 0.4788 & 0.5001 & 0.0036 & 0.0340 & 0.4457 & 0.5002 & 0.0048 & 0.0258 \\
    NPR~\cite{lrrandnpr} $\diamondsuit$ & 0.6088 & 0.6170 & 0.1571 & 0.0185 & 0.5074 & 0.5277 & 0.0612 & 0.0293 & 0.4738 & 0.5204 & 0.0340 & 0.0177 \\
    DetectGPT~\cite{detectgpt} $\diamondsuit$ & 0.6353 & 0.6241 & 0.1719 & 0.0193 & 0.5434 & 0.5515 & 0.0668 & 0.0309 & 0.5079 & 0.5260 & 0.0431 & 0.0239 \\
    Fast-DetectGPT~\cite{fastdetectgpt} & 0.7706 & 0.7193 & 0.2078 & 0.4200 & 0.5727 & 0.5619 & 0.0607 & 0.1238 & 0.5480 & 0.5495 & 0.0525 & 0.1097 \\
    ImBD~\cite{imbd} $\spadesuit$ & 0.8612 & 0.7791 & 0.5599 & 0.4183 & 0.7951 & 0.7199 & 0.4451 & 0.3036 & 0.7694 & 0.6920 & 0.3936 & 0.2868 \\
    \hdashline
    
    \hdashline
    \rowcolor[HTML]{fff5f4}
    \textbf{DetectAnyLLM} (ours) $\spadesuit$ & \textbf{0.9526} & \textbf{0.9059} & \textbf{0.8119} & \textbf{0.7722} & \textbf{0.9316} & \textbf{0.8740} & \textbf{0.7483} & \textbf{0.7779} & \textbf{0.9158} & \textbf{0.8643} & \textbf{0.7320} & \textbf{0.7574} \\
    
    \rowcolor[HTML]{fff5f4}
    \textbf{Imp.} & \red{+65.85\%} & \red{+57.40\%} & \red{+57.25\%} & \red{+60.84\%} & \red{+66.62\%} & \red{+55.02\%} & \red{+54.64\%} & \red{+68.11\%} & \red{+63.49\%} & \red{+55.94\%} & \red{+55.80\%} & \red{+65.98\%} \\
    \hline

    \hline

    \hline
    \end{tabular}
    }
    
    \label{tab:main_results}
\end{table*}
\section{Experiment}
\subsection{Main Results}\label{main_result}

\noindent \textbf{Training settings. }
The scoring model and training data used in DetectAnyLLM are set exactly the same as ~\cite{imbd} to ensure a fair comparison.
Detailed training settings are provided in the Supplementary Material.
The $\gamma$ in DDL is set to 100, and we will discuss how the $\gamma$ affects the performance in \Cref{ablation_study}.

\noindent \textbf{Baselines. }
For a comprehensive comparison, we compare the performance of our method with baseline methods, advanced zero-shot methods, and state-of-the-art training-based methods.
The baseline methods including \textit{Likelihood}~\cite{likelihood}, \textit{Log-Rank}~\cite{logrank}, \textit{LRR}~\cite{lrrandnpr}, and \textit{Entropy}~\cite{entropy}.
The advanced zero-shot methods includes \textit{DetectGPT}~\cite{detectgpt}, \textit{NPR}~\cite{lrrandnpr}, and \textit{Fast-DetectGPT}~\cite{fastdetectgpt}.
The training-based methods includes \textit{RoBERTa series}~\cite{roberta, likelihood} and \textit{ImBD}~\cite{imbd}.

\noindent \textbf{Results on MIRAGE-DIG. }
As the top of \Cref{tab:main_results} shown, DetectAnyLLM achieves substantial performance improvement over all baselines across all metrics and tasks.
Specifically, it delivers AUROC relative gains of up to \red{+64.78\%}$\sim$ \red{+66.71\%}, with MCC improvements reaching up to \red{+56.44\%}.
DetectAnyLLM also maintains robust TPR@5\% across all tasks, outperforming previous training-based SOTA ImBD~\cite{imbd} by large margins (\red{+60.84\%}$\sim$ \red{+69.13\%}).

\noindent \textbf{Results on MIRAGE-SIG. }
As the bottom of~\Cref{tab:main_results} shown, DetectAnyLLM continues to lead in the MIRAGE-SIG subset, reaching AUROC of \red{0.9526}, Balanced Accuracy of \red{0.9059}, and TPR@5\% up to \red{0.7779}, again greatly surpassing all other methods.

The results on MIRAGE highlight DetectAnyLLM's strong generalization capacity and robustness across diverse source LLMs and text-domains, demonstrating the great effectiveness of DDL.

\begin{table}[htbp]
    \centering
    \caption{Results of detection on the previous test sets. $\spadesuit$ indicates a training-based method, whereas $\diamondsuit$ denotes a method that requires multiple model invocations. "Imp." represents Improvement, computed as $(new - old) / (1.0 - old)$.}
    \resizebox{\linewidth}{!}{
    \begin{tabular}{l|ccc}
    \hline

    \hline

    \hline
    \multicolumn{4}{c}{\textbf{ImBD~\cite{imbd} Test Dataset (GPT-4o polished)}}\\
    \hline

    \hline

    \hline
    Methods&XSum~\cite{xsum}&Writing~\cite{writting_prompts} & PubMed~\cite{pubmedqa}\\
    \hline

    \hline
    Likelihood~\cite{likelihood} & 0.4396 & 0.8077 & 0.4596\\
    LogRank~\cite{logrank} & 0.4002 & 0.7694 & 0.4472\\
    Entropy~\cite{entropy} & 0.6122 & 0.2802 & 0.5899\\
    RoBERTa-Base~\cite{roberta} $\spadesuit$ & 0.4921 & 0.4774 & 0.2496 \\
    RoBERTa-Large~\cite{roberta} $\spadesuit$ & 0.4782 & 0.4708 & 0.3089 \\
    LRR~\cite{lrrandnpr} & 0.3095 & 0.6214 & 0.4710 \\
    DNA-GPT~\cite{dna-gpt} $\diamondsuit$ & 0.4974 & 0.7478 & 0.3151 \\
    NPR~\cite{lrrandnpr} $\diamondsuit$ & 0.5065 & 0.8444 & 0.3740 \\
    DetectGPT~\cite{detectgpt} $\diamondsuit$ & 0.6217 & 0.8771 & 0.5612\\
    Fast-DetectGPT~\cite{fastdetectgpt} & 0.6293 & 0.8324 & 0.6175\\
    ImBD~\cite{imbd} $\spadesuit$ & 0.9486 & 0.9468 & 0.7743\\
    \hdashline
    
    \hdashline
    \rowcolor[HTML]{fff5f4}
    \textbf{DetectAnyLLM} (ours) $\spadesuit$ & \textbf{0.9880} & \textbf{0.9671} & \textbf{0.8817} \\
    
    \rowcolor[HTML]{fff5f4}
    \textbf{Imp.} & \red{+80.16\%} & \red{+38.16\%} & \red{+47.59\%} \\
    \hline

    \hline

    \hline
    \end{tabular}
    }
    
    \label{tab:previous_dataset}
\end{table}

\noindent \textbf{Detection on the previous test sets. }
We evaluate the performance of DetectAnyLLM on the three test sets used by ImBD~\cite{imbd}.
As ~\Cref{tab:previous_dataset} shown, DetectAnyLLM consistently outperforms all existing MGTD methods.

Comparing ~\Cref{tab:previous_dataset} and ~\Cref{tab:main_results}, we observe that the baseline methods exhibit great performance degradation on MIRAGE.
Such an observation reveals the limitations of existing test benchmarks in comprehensively evaluating detectors' abilities, underscoring the importance of MIRAGE as a more challenging benchmark.

\noindent \textbf{Efficiency. }
Since DDL performs optimization without relying on a reference model, it achieves substantial improvements in training efficiency compared to \textit{Style Preference Optimization (SPO)}~\cite{imbd}.
DDL demonstrates a \red{+30.12\%} reduction in training time and a \red{+35.90\%} reduction in memory consumption relative to SPO~\cite{imbd}.
Details are provided in Supplementary Material.

\subsection{Ablation Study}\label{ablation_study}
\noindent \textbf{Ablation on parameter $\gamma$. }
As shown in ~\Cref{tab:ablation_gamma}, DDL exhibits strong robustness to the values of $\gamma$.
Comparing to the results in ~\Cref{tab:main_results}, for all selected values of $\gamma$, the DDL-trained detector consistently outperforms all prior state-of-the-art methods in terms of AUROC.
Detailed results, comprehensive analysis, and discussion are provided in Supplementary Material.

\begin{table}[htbp]
    \centering
    \caption{Results of different $\gamma$ in DDL. Metrics with subscript $_t$ correspond to the training set, and subscript $_v$ indicates evaluation on the polish task of MIRAGE-DIG.}
    \resizebox{\linewidth}{!}{
    \begin{tabular}{l|cccccc}
    \hline

    \hline

    \hline
    & $\gamma=10$  & $\gamma=20$ & $\gamma = 50$  & $\gamma=100$  & $\gamma=500$ & $\gamma=10000$\\
    \hline
    AUROC$_t$  &  \textbf{0.9964} & 0.9934 & 0.9883 & 0.9861 & 0.9861 & 0.9861\\
    AUPR$_t$   &  \textbf{0.9965} & 0.9938 & 0.9888 & 0.9833 & 0.9833 & 0.9833\\
    \hdashline
    \rowcolor[HTML]{fff5f4}
    AUROC$_v$  & 0.8692  & 0.9257 & \textbf{0.9347} & 0.9259 & 0.9259 & 0.9259\\
    \rowcolor[HTML]{fff5f4}
    AUPR$_v$   & 0.8735  & 0.9294 & \textbf{0.9458} & 0.9373 & 0.9373 & 0.9373\\
    \hline

    \hline

    \hline
    \end{tabular}
    }
    \label{tab:ablation_gamma}
\end{table}

\noindent \textbf{Ablation on KL-strength $\beta$ in SPO~\cite{imbd}. } We provide comprehensive experiments and confirm our point in~\Cref{redundant KL-regularization}.
For more information, please see Supplementary Material.

\noindent \textbf{Ablation on model. }
We retrain Qwen2-0.5B~\cite{qwen2}, GPT-J-6B~\cite{gpt-j}, and GPT-Neo-2.7B~\cite{gpt-neo} using SPO~\cite{imbd} and DDL.
The models are then evaluated on the \textit{Rewrite} task of MIRAGE-SIG.

As~\Cref{tab:ablation_model_size} shown, the DDL-optimized detector consistently outperforms the SPO~\cite{imbd}-optimized ones across all model sizes, confirming DDL's robustness and adaptability.
It is worth noting that the detector trained using a smaller but more advanced LLM, such as the Qwen2-0.5B model, achieves a better performance.
This shows that the ability of the scoring model largely affects the upper limit of the detector's ability.

\begin{table}[htbp]
    \centering
    \caption{Ablation study on the impact of scoring model. All models are trained on the same data and evaluated on the MIRAGE-SIG \textit{Rewrite} task. Improvement is marked \red{red}.}
    \resizebox{\linewidth}{!}{
    \begin{tabular}{c|lcccc}
    \hline

    \hline

    \hline
    Method & Base Model &  AUROC  &  Accuracy  &  MCC & TPR@5\%\\
    \hline
    \multirow{3}{*}{SPO~\cite{imbd}}&Qwen2-0.5B~\cite{qwen2} & 0.8570 & 0.7816 & 0.5632 & 0.4508\\
    &GPT-Neo-2.7B~\cite{gpt-neo}& 0.7694 & 0.6920 & 0.3936 & 0.2868\\
    &GPT-J-6B~\cite{gpt-j}& 0.8367 & 0.7557 & 0.5155 & 0.4722\\
    \hline
    \multirow{6}{*}{DDL (ours)}&\multirow{2}{*}{Qwen2-0.5B~\cite{qwen2}} & 0.9370 & 0.9071 & 0.8169 & 0.8575\\
    && \red{+55.94\%} & \red{+57.46\%} & \red{+58.08\%} & \red{+74.05\%}\\
    \cdashline{2-6}
    &\multirow{2}{*}{GPT-Neo-2.7B~\cite{gpt-neo}}& 0.9158 & 0.8643 & 0.7320 & 0.7574\\
    && \red{+63.49\%} & \red{+55.94\%} & \red{+55.80\%} & \red{+65.98\%}\\
    \cdashline{2-6}
    &\multirow{2}{*}{GPT-J-6B~\cite{gpt-j}}& 0.8909 & 0.8424 & 0.6878 & 0.6047\\
    && \red{+33.19\%} & \red{+35.49\%} & \red{+35.56\%} & \red{+25.10\%}\\
    \hline

    \hline

    \hline
    \end{tabular}
    }
    \label{tab:ablation_model_size}
\end{table}

\noindent \textbf{Ablation on normalization $\sigma$. }\label{ablation_on_sigma}
As shown in~\Cref{tab:ablation_normalization}, the removal of $\sigma$ leads to a substantial degradation in performance across all metrics, highlighting the importance of $\sigma$ for stable and effective optimization.
Despite this, DDL without normalization still surpasses previous state-of-the-art methods on most metrics reported in~\Cref{tab:main_results}, underscoring the robustness of DDL.
We suggest that the normalization item $\sigma$ helps standardize the output across diverse source LLMs and domains, thereby facilitating more consistent and generalizable learning.

\begin{table}[htbp]
    \centering
    \caption{Ablated result of $\sigma$. ``norm.'' means ``normalization''. ``w/'' means ``with'' and ``w/o'' means ``without''. Scoring model: GPT-Neo-2.7B~\cite{gpt-neo}. Benchmark: MIRAGE-DIG-polish.}
    \begin{tabular}{l|cccc}
    \hline

    \hline

    \hline
         &  AUROC  &  Accuracy  &  MCC & TPR@5\%\\
    \hline
    DDL$_{\textbf{w/o}\enspace norm.}$ & 0.8499 & 0.7759 & 0.5563 & 0.5232\\
    \hdashline
    \rowcolor[HTML]{fff5f4}
    DDL$_{\textbf{w/}\enspace norm.}$& \textbf{0.9297} & \textbf{0.8732} & \textbf{0.7487} & \textbf{0.7756}\\
    \hline

    \hline

    \hline
    \end{tabular}
    \label{tab:ablation_normalization}
\end{table}
\section{Conclusion}
In this study, we have introduced a novel optimization strategy, \textit{Direct Discrepancy Learning (DDL)}, and developed a unified detection framework named \textit{DetectAnyLLM}.
Our approach enables the scoring model to acquire task-oriented knowledge by directly leveraging discrepancy signals and achieves high-precision detection through a technique we call \textit{reference clustering}.
We also proposed \textit{MIRAGE}, a comprehensive benchmark dataset that spans a wide range of text domains, the most advanced LLMs, and generation tasks.
To thoroughly evaluate detector performance, we assessed DetectAnyLLM and existing MGTD methods under two settings: \textit{Disjoint-Input Generation} and \textit{Shared-Input Generation}.
Experimental results on both MIRAGE and previously established test sets demonstrate that DetectAnyLLM significantly outperforms existing MGTD methods, establishing a new state-of-the-art in this domain.

\begin{acks}
This work was supported in part by the National Natural Science Foundation of China (62306153, 62225604), the Natural Science Foundation of Tianjin, China (24JCJQJC00020), the Young Elite Scientists Sponsorship Program by CAST (YESS20240686), the Fundamental Research Funds for the Central Universities (Nankai University, 070-63243143),  and Shenzhen Science and Technology Program (JCYJ20240813114237048). 

The computational devices are partly supported by the Supercomputing Center of Nankai University (NKSC).
\end{acks}

\bibliographystyle{sec/ACM-Reference-Format}
\balance
\bibliography{main}

\clearpage
\appendix
\section{More details on DDL}
\subsection{Derivation of DPO}
\textit{Direct Preference Learning (DPO)}~\cite{dpo} is derived from the optimization objective of \textit{Proximal Policy Optimization (PPO)}~\cite{ppo}, which is formulated as:
\begin{equation}
    \mathop{max}_\theta \mathbb{E}_{x\sim f_\theta(x)}[r(x)] - \beta\mathbb{D}_{KL}[(f_\theta(x) \mid \mid f_{ref}(x)],
    \label{dpo_1}
\end{equation}
where $x$ is a text sampled from the scoring model $f_\theta$'s distribution, and $r$ is a reward function that can judge whether this sample is bad or good.
We can obtain DPO's optimization objective by analyzing and re-parameterizing this optimization objective.
The $f_{ref}$ stands for the reference model, usually the original $f_\theta$.
DPO~\cite{dpo} starts with explicit solution $f_\theta=p_r$ of Eq.~\eqref{dpo_1}:
\begin{equation}
    p_r(x) = \frac{1}{Z(x)}f_{ref}(x)\exp(\frac{1}{\beta}r(x)),
    \label{dpo_2}
\end{equation}
where:
\begin{equation}
    Z(x) = \sum_{x}f_{ref}(x)\exp (\frac{1}{\beta}r(x)).
    \label{dpo_3}
\end{equation}
Furthermore, we can reparameterize $r$ as:
\begin{equation}
    r(x) = \beta\log \frac{p_r(x)}{f_{ref}(x)} + \beta\log Z(x),
    \label{dpo_4}
\end{equation}
where, $p_r$ is the best solution of Eq.~\eqref{dpo_1}, which we want $f_\theta$ to become.
Now, if we introduce the Bradley-Terry model to present the model's preference $L$ between HWT $x_h$ and MGT $x_m$, we can get:
\begin{equation}
\begin{aligned}
    L(x_m \succ x_h) &= \sigma(r(x_m) - r(x_h))\\
    &=\sigma(\beta\log\frac{p_r(x_m)}{f_{ref}(x_m)} - \beta\log\frac{p_r(x_h)}{f_{ref}(x_h)}),
\end{aligned}
\label{dpo_5}
\end{equation}
where we surprisingly reduced the partition function $Z(x)$.
By replacing $p_r$ with $f_\theta$ and utilizing Maximum-Likelihood Estimation to Eq.~\eqref{dpo_5}, we can finally present the DPO~\cite{dpo} optimize goal :
\begin{equation}
    \mathop{max}_\theta \mathbb{E}_{x_m, x_h \sim \mathop{D}}[\log \sigma(\beta\log\frac{f_\theta(x_m)}{f_{ref}(x_m)}-\beta\log\frac{f_\theta(x_h)}{f_{ref}(x_h)})],
    \label{dpo_6}
\end{equation}
where $x_m$ denotes MGT and $x_h$ stands for HWT.
$f_{ref}$ is a reference model, usually the original $f_\theta$.

\section{More details on MIRAGE}
\subsection{More details on Data Source}
\noindent \textbf{Time Bound. }
To ensure that all sampled texts are human-written and free from contamination by LLM-generated content, most of the source datasets used were constructed prior to 2021.
For datasets containing data collected after 2021, we cleaned the data denoted collected after 2021 in these datasets to ensure the purity and authenticity of the human-written source material.

\noindent \textbf{Source Domains and Datasets. }
MIRAGE encompasses a diverse range of text domains, including \textit{Academic}, \textit{E-Mail}, \textit{Website}, \textit{News}, and \textit{Comment}.
The mapping between these domains and the corresponding source datasets is summarized in~\Cref{tab:source_datasets}.

\begin{table}[h]
    \centering
    \renewcommand{\arraystretch}{1.25}
    \caption{Source datasets for each domain. }
    \resizebox{0.985\linewidth}{!}{
    \begin{tabular}{l|c}
    \hline
    
    \hline
    
    \hline
    \textbf{Domains} &  \textbf{Datasets} \\
    \hline
    Academic  & BigPatent~\cite{bigpatent}, NeurIPS, ArXiv, PubMed-Abstracts~\cite{pubmedqa}\\
    eMail & Enron-Emails\\
    Website & OpenWebText\\
    News & CNNDailyMails, XSum~\cite{xsum}, XLSum~\cite{xlsum}\\
    Comment &  Amazon-Review\\
    \hline

    \hline
    
    \hline
    \end{tabular}
    }
    \label{tab:source_datasets}
\end{table}

Additionally, we implement domain-specific pre-processing: extracting only abstracts from academic publications (NeurIPS and ArXiv) and isolating message content from email communications (Enron-Emails dataset).

\noindent \textbf{Domain Composition. }
As the amount of data varies across domains, the text domains are not treated equally in terms of quantity.
However, for both DIG and SIG, the proportion of texts from each domain that each LLM is required to generate or revise remains fixed.
The detailed domain distribution is shown in~\Cref{tab:domain_composition}.

\begin{table}[htbp]
    \centering
    \renewcommand{\arraystretch}{1.25}
    \caption{Text domain composition that each LLM is required to perform generation task in both DIG and SIG.}
    \begin{tabular}{ccccc|c}
    \hline

    \hline

    \hline
    Academic& Mail & Website & News & Comment & \textbf{Total}  \\
    \hline
    300 & 100 & 200 & 200 & 200 & 1000 \\
    \hline

    \hline

    \hline
    \end{tabular}
    \label{tab:domain_composition}
\end{table}

\noindent \textbf{Statistic Result. }
\Cref{tab:statistic} presents the overall statistics of the MIRAGE dataset across the two task settings: Disjoint-Input Generation and Shared-Input Generation.
For each setting, the dataset includes three task types—\textit{Generate} (Gen.), \textit{Polish} (Pol.), and \textit{Rewrite} (Rew.).
The number of instances is balanced across both settings, with each task type containing approximately 14,000 to 16,000 samples.
This balanced distribution ensures that the dataset supports a comprehensive evaluation of LLMs across different generation and revision scenarios.

\begin{table}[htbp]
    \centering
    \renewcommand{\arraystretch}{1.1}
    \caption{Statistic result of MIRAGE.}
    \resizebox{0.985\linewidth}{!}{
    \begin{tabular}{l|ccc|ccc}
    \hline

    \hline

    \hline
            & \multicolumn{3}{c|}{Disjoint-Input Generation} & \multicolumn{3}{c}{Shared-Input Generation} \\
    Tasks& Gen. & Pol. & Rew. & Gen. & Pol. & Rew. \\
    \hline
    Count& 16412 & 14776 & 15735 & 16388 & 14776 & 15751 \\
    \hline

    \hline

    \hline
    \end{tabular}
    }
    \label{tab:statistic}
\end{table}

\subsection{Source Generator LLMs}
The source LLMs used for data generation in MIRAGE are listed in~\Cref{tab:src_models}.
In total, MIRAGE samples machine-generated texts (MGT) using 13 powerful commercial LLMs and 4 advanced open-source LLMs.
This selection reflects a strong emphasis on evaluating detection performance in real-world applicant scenarios, while still maintaining attention to the open-source LLM landscape.

\begin{table}[htbp]
    \centering
    \renewcommand{\arraystretch}{1.25}
    \caption{Commercial LLMs are highlighted in \textbf{bold}.}
    \resizebox{0.985\linewidth}{!}{
    \begin{tabular}{c|c}
    \hline

    \hline

    \hline
    Series & Models \\
    \hline
    GPT&  \textbf{GPT-4o}~\cite{gpt4o}, \textbf{GPT-o3-mini}~\cite{gpto1}, \textbf{GPT-4o-mini}~\cite{gpt4o}\\
    Claude& \textbf{Claude-3.5-Haiku}, \textbf{Claude-3.7-sonnet}~\cite{claude} \\
    DeepSeek& \textbf{DeepSeek-V3}~\cite{deepseekv3}, \textbf{DeepSeek-R1}~\cite{deepseekr1}\\
    Gemini& \textbf{Gemini-2.0-Flash}, \textbf{Gemini-2.0-Flash-Lite}~\cite{gemini} \\
    Qwen& Qwen-2.5-7B~\cite{qwen2.5}, Qwen-2.5-7B-R1-Distill~\cite{deepseekr1}, \textbf{QwQ-Plus}\\
    LlaMa& LlaMA-3.1-8B~\cite{llama3}, LlaMa-3.1-8B-R1-Distill~\cite{deepseekr1}\\
    Grok& \textbf{Grok2} \\
    Moonshot & \textbf{Moonshot-v1} \\
    Doubao& \textbf{Doubao-1.5-pro-32k} \\
    \hline

    \hline

    \hline
    \end{tabular}
    }
    \label{tab:src_models}
\end{table}

\subsection{Prompts for Generation Tasks}
The system prompts used for all three generation tasks are the same, specifically, ``You are a professional writing assistant who can write high-quality, coherent, and engaging articles. "
We add a style control signal to the user prompt, in order to perform data augmentation, thus promoting the robustness of our benchmark.

\noindent \textbf{Style Control. }
The style control signal is directly added to the user prompt, as ``in a $<$style$>$ style".
The $<$style$>$ is randomly chosen from a prepared style list, detail as shown in ~\Cref{tab:style_list}
\begin{table}[htbp]
    \centering
    \caption{Detail style list.}
    \begin{tabular}{c}
    \hline

    \hline

    \hline
    Style List \\
    \hline
    formal, oral, academic, literary, critical, narrative, \\
    descriptive, lyric, objective, subjective, original,\\
    casual, expository, argumentative, journalistic, poetic\\
    \hline

    \hline

    \hline
    \end{tabular}
    \label{tab:style_list}
\end{table}

\noindent \textbf{Prompt for Generate. }
``Write an article about 150 words in a $<$style$>$ style starting exactly with: $<$original$>$".
The $<$original$>$ is the first 30 tokens of a HWT.

\noindent \textbf{Prompt for Polish. }
``Polish the following text in a {style} style without missing any original details. Ensure that the length of the polished text is similar to the original text. Directly output your polished text. Here is the original text: {original}"
The $<$original$>$ is a complete HWT.

\noindent \textbf{Prompt for Rewrite. }
``Paraphrase the following text in a {style} style without missing any original details. Ensure that the length of the paraphrased text is similar to the original text. Directly output your paraphrased text. Here is the original text: {original}"
The $<$original$>$ is a complete HWT.

\begin{table*}[t]
    \centering
    \caption{Detail Results of different $\beta$ in SPO~\cite{imbd}. Metrics with subscript $_t$ correspond to the training set, and subscript $_v$ indicates evaluation on the polish task of MIRAGE-DIG. Avg.D(*) denotes the average discrepancy of *, where x$_h$ stands for human-written text and x$_m$ stands for machine-generated text. $\Delta$D denotes the distance between Avg.D(x$_h$) and Avg.D(x$_m$), a higher $\Delta$D is commonly better for discriminating between x$_h$ and x$_m$.}
    \resizebox{\linewidth}{!}{
    \begin{tabular}{l|ccccccccccccccc}
    \hline

    \hline

    \hline
    $\beta$& $0.05$ & $0.10$ & $0.15$ & $0.20$ & $0.25$ & $0.30$ & $0.35$ & $0.40$ & $0.45$ & $0.50$ & $0.60$ & $0.70$ & $0.80$ & $0.90$ & $0.95$\\
    \hline
    AUROC$_t$        & \textbf{0.9490} & 0.9192 & 0.9088 & 0.9009 & 0.8945 & 0.8920 & 0.8888 & 0.8871 & 0.8821 & 0.8786 & 0.8742 & 0.8700 & 0.8622 & 0.8554 & 0.8542\\
    AUPR$_t$         & \textbf{0.9566} & 0.9277 & 0.9155 & 0.9058 & 0.8982 & 0.8966 & 0.8935 & 0.8934 & 0.8897 & 0.8869 & 0.8846 & 0.8805 & 0.8741 & 0.8695 & 0.8689\\
    Avg.D(x$_h$)$_t$ & -29.14 & -13.50 & -8.68  & -5.72  & -4.47  & -3.57  & -2.92  & -2.36  & -2.06  & -2.05  & -1.54  & -1.51  & -1.37  & -0.98  & -0.93 \\
    Avg.D(x$_m$)$_t$ & -11.55 & -6.56  & -3.88  & -2.11  & -1.37  & -0.79  & -0.39  & -0.01  &  0.17  &  0.14  &  0.50  &  0.50  &  0.53  &  0.83  &  0.86 \\
    $\Delta$D        &  17.59 &  6.94  &  4.8   &  3.61  &  3.10  &  2.78  &  2.53  &  2.35  &  2.23  &  2.19  &  2.04  &  2.01  &  1.90  &  1.81  &  1.79 \\
    \hdashline
    \rowcolor[HTML]{fff5f4}
    AUROC$_v$        & \textbf{0.7888} & 0.7273 & 0.7045 & 0.6669 & 0.6801 & 0.6786 & 0.6745 & 0.6625 & 0.6688 & 0.6672 & 0.6520 & 0.6644 & 0.6499 & 0.6315 & 0.6477 \\
    \rowcolor[HTML]{fff5f4}
    AUPR$_v$         & \textbf{0.7756} & 0.7122 & 0.6910 & 0.6602 & 0.6783 & 0.6757 & 0.6764 & 0.6663 & 0.6778 & 0.6719 & 0.6597 & 0.6723 & 0.6578 & 0.6365 & 0.6566 \\
    \rowcolor[HTML]{fff5f4}
    Avg.D(x$_h$)$_v$ & -26.28 & -17.65 & -11.50 & -7.30  & -5.30  & -3.78  & -2.67  & -1.82  & -1.33  & -1.05  & -0.47  & -0.45  & -0.26  &  0.32  &  0.34 \\
    \rowcolor[HTML]{fff5f4}
    Avg.D(x$_m$)$_v$ & -18.91 & -14.53 & -9.94  & -6.27  & -4.51  & -3.11  & -2.06  & -1.20  & -0.74  & -0.51  & -0.00  &  0.00  &  0.10  &  0.78  &  0.82 \\
    \rowcolor[HTML]{fff5f4}
    $\Delta$D        &  7.37  &  3.12  &  1.56  &  1.03  &  0.79  &  0.67  &  0.61  &  0.62  &  0.59  &  0.54  &  0.47  &  0.45  &  0.36  &  0.46  &  0.48 \\
    \hline

    \hline

    \hline
    \end{tabular}
    }
    \label{tab:detail_ablation_beta}
\end{table*}

\subsection{More Details on  Evaluation Metrics}
Consistent with prior work~\cite{fastdetectgpt, imbd}, we adopt \textit{the Area Under the Receiver Operating Characteristic Curve (AUROC)} as the primary evaluation metric for assessing the performance of the MGT Detector.
While AUROC provides a threshold-independent measure of classification capability, it does not necessarily reflect the detector’s effectiveness at specific operating points, which are often critical in real-world deployments.

To address this limitation, we incorporate \textit{TPR at a 5\% false positive rate (TPR@5\%)} as an important supplementary metric.
TPR@5\% reflects the detector’s sensitivity when operating under a strict false positive constraint, which is especially important for applications demanding high precision.

Furthermore, considering the MIRAGE-SIG is a class-imbalanced dataset, we additionally report the \textit{Matthews Correlation Coefficient (MCC)} and \textit{Balanced Accuracy} to provide a more comprehensive evaluation.
MCC captures the overall quality of binary classifications by considering all four elements of the confusion matrix, making it particularly informative under class imbalance.
Balanced Accuracy, used in place of standard accuracy, is computed as the arithmetic mean of the true positive rate and true negative rate, making it better suited for evaluating performance on imbalanced datasets.

Together, this diverse set of metrics provides a comprehensive assessment of the detector's performance, ensuring that the evaluation reflects both theoretical completeness and real-world applicability.

\section{More details on Experiment}
\subsection{Experiment Setup}
\noindent \textbf{Device. }
All of our experiments are conducted in Linux 4.18.0(CentOS 7), using a single NVIDIA A40 GPU with 48GB GPU memory.
The Python version is 3.10.16, the PyTorch version is 2.5.1, the Transformers version is 4.47.1, and the Datasets version is 3.2.0.

\noindent \textbf{Training Dataset. }
We train DetectAnyLLM in the dataset used in ImBD~\cite{imbd}, specifically, 500 pairs of HWT-MGT data, where MGT is machine-polished text created by GPT-3.5-Turbo.

\noindent \textbf{LoRA Config. }
Following the settings of ImBD~\cite{imbd}, we adopt a LoRA configuration specifically designed for causal language modeling, with a rank of 8, a LoRA alpha of 32, and a dropout rate of 0.1.

\noindent \textbf{Settings for Reproducing ImBD. }
We reproduced ImBD~\cite{imbd} for comparative evaluation, following the original training configuration described in the paper.
Specifically, we set the learning rate to 0.0001 and used a beta coefficient of 0.05.
The only modification made was increasing the number of training epochs from 2 (as reported in the original paper) to 5, in order to ensure full convergence of the model.
Throughout the training process, we monitored the model's performance on the validation set to prevent overfitting and to verify that the reproduced ImBD model maintained comparable performance to the original.

\noindent \textbf{Settings for Training DetectAnyLLM. }
We train DetectAnyLLM using the exact same hyperparameters as those used in our reproduction of ImBD~\cite{imbd}, including a learning rate of 0.0001 and 5 training epochs.
For the optimization objective in \textit{Direct Discrepancy Learning (DDL)}, we set the hyperparameter $\gamma$ to 100.
That is because increasing $\gamma$ beyond this value did not lead to further improvements in performance, suggesting that the model had fully converged.
Moreover, since the model's performance remains stable for larger values of $\gamma$, this setting also ensures compatibility with varying training environments, as the optimal value of $\gamma$ is unknown, we simply choose a sufficiently large value that provides a safe and generalizable configuration.

\subsection{Empirical Validation of Redundant KL-Regularization}
To evaluate the impact of the KL-regularization term in DPO-style training, we conduct ablation studies on a wide range of $\beta$ values, which directly control the strength of the implicit KL constraint. According to the formulation in Section 3.2, a larger $\beta$ enforces stronger alignment between the scoring model $f_\theta$ and the reference model $f_{ref}$, effectively constraining the learning objective toward distributional conformity rather than task-specific discriminability. 

~\Cref{tab:detail_ablation_beta} provides compelling empirical evidence supporting our hypothesis that KL-regularization can be redundant—or even detrimental—for the MGTD task. As $\beta$ increases, we observe a consistent and significant degradation in detection performance across all evaluation metrics. For instance, the AUROC and AUPR on the training set drop from \textbf{0.9490}/\textbf{0.9566} at $\beta=0.05$ to 0.8542/0.8689 at $\beta=0.95$. Similar trends are observed on the validation set, where AUROC decreases from 0.7888 to 0.6477, and AUPR from 0.7756 to 0.6566.

At low $\beta$ values, the discrepancy between D(x$_h$) and D(x$_m$) is substantial (e.g., $17.59$ on the training set at $\beta=0.05$), which allows the scoring model to effectively differentiate between natural and perturbed sequences. As $\beta$ increases, this margin rapidly collapses—dropping below $3.0$ by $\beta=0.30$, and approaching near-zero at higher values. The validation set exhibits a parallel pattern: the discrepancy narrows from $7.37$ at $\beta=0.05$ to merely $0.48$ at $\beta=0.95$.

These results indicate that strong KL-regularization impairs the model's ability to learn task-oriented discriminative signals from training data. Instead of becoming a better detector, the scoring model is constrained to behave like a generic language model, limiting its effectiveness in distinguishing machine-generated text from human-written. This validates our theoretical intuition: while the KL term may serve to preserve internal knowledge in generic preference modeling tasks, it is counterproductive in MGTD.

\subsection{More Details on Main Results}
\noindent \textbf{Efficiency Improvement. }
When using the scoring model $f_\theta$ as the sampling model $q_\phi$, as discussed in Section 3.2, DDL eliminates the need to load a separate reference model during training, unlike SPO~\cite{imbd}.
This design enables DDL to train with a single model, leading to notable improvements in training efficiency.
\Cref{tab:efficiency} presents a detailed comparison of training time and memory usage between SPO~\cite{imbd} and DDL.

SPO~\cite{imbd} requires loading two large models simultaneously during training, resulting in high memory demands—specifically, 31.45GB for training with GPT-J-6B~\cite{gpt-j}. This exceeds the capacity of many commonly available GPUs.
In contrast, DDL only requires 20.16GB of memory, making it feasible to train on widely accessible GPUs.

\begin{table}[t]
    \centering
    \caption{Training time cost comparison between DDL and SPO~\cite{imbd}. The results are tested in ImBD~\cite{imbd}'s training dataset. Device: single NVIDIA A40. Model: GPT-J-6B~\cite{gpt-j}. ``Imp." represents Improvement, computed as -(new - old) / (old).}
    \begin{tabular}{l|ccc}
    \hline

    \hline

    \hline
    Optim.          &  Batch Size & Time Cost/Epoch  &  Memory Usage\\
    \hline
    SPO~\cite{imbd} & 1 & 166s & 31.45GB\\
    \hdashline
    \rowcolor[HTML]{F4F7FE}
    DDL(ours)& 1 & \textbf{116s} & \textbf{20.16GB}\\
    \rowcolor[HTML]{F4F7FE}
    Imp. & -- & \red{+30.12\%} & \red{+35.90\%}\\
    \hline

    \hline

    \hline
    \end{tabular}
    \label{tab:efficiency}
\end{table}

\begin{table*}[t]
    \centering
    \caption{Detail Results of different $\gamma$ in DDL. Metrics with subscript $_t$ correspond to the training set, and subscript $_v$ indicates evaluation on the polish task of MIRAGE-DIG. Avg.D(*) denotes the average discrepancy of *, where x$_h$ stands for human-written text and x$_m$ stands for machine-generated text. }
    \resizebox{\linewidth}{!}{
    \begin{tabular}{l|ccccccccccccccc}
    \hline

    \hline

    \hline
    & $\gamma=1$ & $\gamma=2$ & $\gamma=5$ & $\gamma=10$ & $\gamma=20$ & $\gamma = 30$ & $\gamma=40$ & $\gamma=50$ & $\gamma=60$ & $\gamma=70$ & $\gamma=80$ & $\gamma = 90$ & $\gamma=100$ & $\gamma=500$ & $\gamma=10000$\\
    \hline
    AUROC$_t$  &  0.9501 &  0.9910 & \textbf{0.9983} & 0.9964 & 0.9934 & 0.9900 & 0.9886 & 0.9883 & 0.9880 & 0.9879 & 0.9861 & 0.9861 & 0.9861 & 0.9861 & 0.9861\\
    AUPR$_t$   &  0.9379 &  0.9910 & \textbf{0.9983} & 0.9965 & 0.9938 & 0.9911 & 0.9865 & 0.9888 & 0.9852 & 0.9852 & 0.9833 & 0.9833 & 0.9833 & 0.9833 & 0.9833\\
    Avg.D(x$_h$)$_t$ & 0.07 & 0.14 &  0.20  &  0.27   &  0.54  &  0.80  &  1.08  &  1.45  &  1.55  &  1.73  &  1.91  &  1.91  &  1.91  &  1.91  &  1.91  \\
    Avg.D(x$_m$)$_t$ & 0.95 & 1.88 &  4.86  &  8.92   &  17.24 &  24.64 &  31.82 &  37.93 &  40.81 &  41.92 &  42.12 &  42.12 &  42.12 &  42.12 &  42.12 \\
    \hdashline
    \rowcolor[HTML]{fff5f4}
    AUROC$_v$  & 0.5481 &  0.6000  & 0.7833 & 0.8692 & 0.9257 & 0.9360 & \textbf{0.9377} & 0.9347 & 0.9251 & 0.9270 & 0.9259 & 0.9259 & 0.9259 & 0.9259 & 0.9259\\
    \rowcolor[HTML]{fff5f4}
    AUPR$_v$   & 0.5206 &  0.5562  & 0.7452 & 0.8735 & 0.9294 & \textbf{0.9472} & 0.9461 & 0.9458 & 0.9401 & 0.9382 & 0.9373 & 0.9373 & 0.9373 & 0.9373 & 0.9373\\
    \rowcolor[HTML]{fff5f4}
    Avg.D(x$_h$)$_v$ & 1.1 & 1.24  &  1.76  &  0.84  &  3.11  &   3.37  &  4.66  &  4.62  &  4.72  &  5.36  &  5.23  &  5.23  &  5.23  &  5.23  &  5.23  \\
    \rowcolor[HTML]{fff5f4}
    Avg.D(x$_m$)$_v$ & 1.36 & 1.85 &  4.86  &  7.11  &  16.09 &   24.22 &  32.96 &  34.86 &  36.75 &  39.86 &  39.43 &  39.43 &  39.43 &  39.43 &  39.43 \\
    \hline

    \hline

    \hline
    \end{tabular}
    }
    \label{tab:detail_ablation_gamma}
\end{table*}

\subsection{Discussion of $\gamma$ in DDL}
\noindent \textbf{Detail Look of $\gamma$'s effect. }
~\Cref{tab:detail_ablation_gamma} shown, although there are clear performance peaks at specific values (e.g., $\gamma=5$ for the training set and $\gamma=30$–$40$ for the validation set), the metrics remain consistently high across a wide range of $\gamma$ values. For instance, even when $\gamma$ increases from 10 to 10000, AUROC$_t$ and AUPR$_t$ only experience a minor drop (from around 0.9964 to 0.9861), and AUROC$_v$ and AUPR$_v$ stay stable after reaching their respective optima.

Meanwhile, AUROC$_v$ and AUPR$_v$ continue to improve up to $\gamma=30$–$\gamma=40$, reaching their peaks at $\gamma=30$ (AUPR$_v$ = 0.9472) and $\gamma=40$ (AUROC$_v$ = 0.9377), after which they plateau.

\noindent \textbf{DDL's Robustness on $\gamma$. }
While the average discrepancies and training metrics vary as $\gamma$ increases, the evaluation performance (AUROC$_v$ and AUPR$_v$) remains relatively stable across a broad range—from $\gamma=30$ to $\gamma=10000$. For example, AUROC$_v$ fluctuates within a narrow band around 0.93, and AUPR$_v$ stays above 0.93 even when $\gamma$ changes by orders of magnitude. This indicates that although $\gamma$ influences how positive the discrepancy of x$_m$ should be, the model's downstream generalization ability is not overly sensitive to its exact value.

This plateauing effect indicates that once $\gamma$ surpasses a moderate threshold, the method maintains strong performance without being overly sensitive to further increases. Moreover, the saturation of Avg.D(x$_h$) and Avg.D(x$_m$) after a certain point suggests the model's behavior becomes consistent, avoiding unstable shifts.
This is a desirable property in practical scenarios, where tuning hyperparameters like $\gamma$ might be challenging or resource-intensive.

\noindent \textbf{Explanation of $\gamma$'s Effectiveness. }
The robustness of our method to the hyperparameter $\gamma$ arises naturally from the design of the \textit{Direct Discrepancy Learning (DDL)}. In DDL, $\gamma$ serves as a margin that guides the optimization: it encourages the model to keep the discrepancy score D(x$_m$) of MGT close to $\gamma$, while minimizing the discrepancy score D(x$_h$) for HWT toward zero. This setup constructs an explicit separation objective in the discrepancy space between HWT and MGT.

\textit{We realized the goal of \textbf{learning to be a detector rather than another language model} by introducing such explicit separation objective. }

A small $\gamma$ (e.g., 1–5) still may encourage separation, however, may not provide enough margin, leading to overlap between HWT and MGT discrepancies. As $\gamma$ increases, the model is encouraged to push D(x$_m$) further away from zero, thus improving distinguishability and enhancing performance—especially noticeable in the rise of AUROC/AUPR metrics from $\gamma=1$ to $\gamma=5$ and beyond.

\noindent \textbf{Explanation of Performance Plateau. }
Once $\gamma$ exceeds a certain threshold (e.g., $\gamma \geq 30$), we observe that performance metrics (both AUROC$_t$ and AUROC$_v$) are saturate. This indicates that the discrepancy between HWT and MGT has reached a sufficient margin: D(x$_h$) is consistently near 0 and D(x$_m$) is already large enough. Increasing $\gamma$ further only increases the target discrepancy for $x_m$ without changing the classification boundary. The model thus stabilizes, as it can no longer extract additional useful separation from a larger $\gamma$. This explains the observed robustness—DDL achieves effective separation and remains performance over a wide range of $\gamma$ values.

\noindent \textbf{Reason of Setting $\gamma$ to 100. }
We choose $\gamma=100$ in our main results for two key reasons. First, as shown in the ablation results, performance has already plateaued by $\gamma=100$, meaning this setting offers high performance while avoiding sensitivity to further hyperparameter tuning—making it a practical and reliable choice in real-world applications. Second, a higher $\gamma$ ensures a clear and consistent discrepancy margin, improving interpretability and stability across diverse datasets or models. In deployment scenarios, where extensive hyperparameter sweeps may be infeasible, such robustness is critical.

\noindent \textbf{Summary. }
As shown in ~\Cref{tab:detail_ablation_gamma}, there exists a threshold value $t_h$ such that the performance remains stable for all $\gamma \geq t_h$.
In contrast, setting $\gamma$ too small can lead to significantly degraded performance, while increasing $\gamma$ beyond $t_h$ does not cause any sharp decline.
Therefore, we recommend setting $\gamma$ to a relatively large value, since in real-world applications the optimal value is typically unknown.

\subsection{Detection Results on specific LLM}
We expand the main results in Section 5.1 in the specific LLM level to obtain the specific detection capabilities of different methods on texts generated by specific LLMs.

\noindent \textbf{Results. }
As shown in the following tables, DetectAnyLLM achieves consistently strong performance across all metrics, domains, tasks, and source LLMs.
On the Polish and Rewrite tasks, it outperforms previous state-of-the-art methods by an average margin of nearly 70\%.
In certain settings involving text generated by specific LLMs, FastDetectGPT\cite{fastdetectgpt} and ImBD\cite{imbd} slightly outperform DetectAnyLLM. We attribute this to the relative simplicity of the Generate task, which allows earlier methods to perform competitively.
Even in these cases, DetectAnyLLM trails by no more than $10^{-2}$ AUROC, suggesting that may have reached saturation on this task.

As shown in ~\Cref{tab:gpt4o_gpt4omini} and ~\Cref{tab:claude3.5haiku_claude3.7sonnet}, the AUROC on DIG text polished by Claude-3.5-Haiku reaches \red{0.9903}, while that of Claude-3.7-Sonnet is \red{0.9096}. Similarly, the AUROC on SIG text rewritten by GPT-4o-mini reaches \red{0.9176}, compared to \red{0.8697} for GPT-4o.
These results indicate that detecting text from smaller LLMs is generally easier than from their larger counterparts.

\begin{table*}[h]
    \centering
    \caption{Generator: GPT-4o, GPT-4o-mini. "Imp.": Improvement over previous SOTA, computed as $(new - old) / (1.0 - old)$.}
    \resizebox{\linewidth}{!}{
\begin{tabular}{l|cccc|cccc|cccc}
    \hline

    \hline

    \hline
    \multicolumn{13}{c}{\textbf{MIRAGE-DIG, GPT-4o}}\\
    \hline

    \hline

    \hline
    \multirow{2}{*}{Methods}&\multicolumn{4}{c|}{Generate}&\multicolumn{4}{c|}{Polish}&\multicolumn{4}{c}{Rewrite} \\
    &  AUROC  &  Accuracy  &  MCC  &  TPR@5\%  &  AUROC  &  Accuracy  &  MCC  &  TPR@5\%  &  AUROC  &  Accuracy  &  MCC  &  TPR@5\%  \\
    \hline

    \hline
    Likelihood~\cite{likelihood} & 0.5021 & 0.5476 & 0.1117 & 0.0040 & 0.4732 & 0.5052 & 0.0227 & 0.0216 & 0.4281 & 0.5000 & 0.0000 & 0.0134 \\
    LogRank~\cite{logrank} & 0.5053 & 0.5406 & 0.1011 & 0.0050 & 0.4638 & 0.5005 & 0.0227 & 0.0216 & 0.4191 & 0.5000 & 0.0000 & 0.0113 \\
    Entropy~\cite{entropy} & 0.6414 & 0.6259 & 0.2976 & 0.1103 & 0.5358 & 0.5402 & 0.1002 & 0.0670 & 0.5733 & 0.5520 & 0.1953 & 0.1236 \\
    RoBERTa-Base~\cite{roberta} $\spadesuit$ & 0.4950 & 0.5206 & 0.0857 & 0.0843 & 0.5013 & 0.5036 & 0.0259 & 0.0474 & 0.5244 & 0.5335 & 0.0767 & 0.0639 \\
    RoBERTa-Large~\cite{roberta} $\spadesuit$ & 0.4478 & 0.5005 & 0.0224 & 0.0361 & 0.5715 & 0.5603 & 0.1207 & 0.0732 & 0.6244 & 0.5896 & 0.1849 & 0.0865 \\
\hline

    \hline

    \hline
    \end{tabular}
    }
    \label{tab:gpt4o_gpt4omini}
\end{table*}
\begin{table*}[h]
    \centering
    \caption{Generator: GPT-o3-mini, Moonshot-v1. "Imp.": Improvement over previous SOTA, computed as $(new - old) / (1.0 - old)$.}
    \resizebox{\linewidth}{!}{
    \begin{tabular}{l|cccc|cccc|cccc}
    \hline

    \hline

    \hline
    \multicolumn{13}{c}{\textbf{MIRAGE-DIG, GPT-o3-mini}}\\
    \hline

    \hline

    \hline
    \multirow{2}{*}{Methods}&\multicolumn{4}{c|}{Generate}&\multicolumn{4}{c|}{Polish}&\multicolumn{4}{c}{Rewrite} \\
    &  AUROC  &  Accuracy  &  MCC  &  TPR@5\%  &  AUROC  &  Accuracy  &  MCC  &  TPR@5\%  &  AUROC  &  Accuracy  &  MCC  &  TPR@5\%  \\
    \hline

    \hline
    Likelihood~\cite{likelihood} & 0.2513 & 0.5015 & 0.0317 & 0.0030 & 0.4473 & 0.5022 & 0.0331 & 0.0164 & 0.4064 & 0.5005 & 0.0232 & 0.0086 \\
    LogRank~\cite{logrank} & 0.2465 & 0.5010 & 0.0317 & 0.0030 & 0.4319 & 0.5005 & 0.0135 & 0.0120 & 0.3951 & 0.5000 & 0.0000 & 0.0065 \\
    Entropy~\cite{entropy} & 0.7324 & 0.6819 & 0.3674 & 0.1573 & 0.5222 & 0.5377 & 0.1187 & 0.0350 & 0.5584 & 0.5500 & 0.1343 & 0.1000 \\
    RoBERTa-Base~\cite{roberta} $\spadesuit$ & 0.4050 & 0.5000 & 0.0000 & 0.0080 & 0.4395 & 0.5000 & 0.0000 & 0.0175 & 0.4779 & 0.5005 & 0.0056 & 0.0376 \\
    RoBERTa-Large~\cite{roberta} $\spadesuit$ & 0.4319 & 0.5005 & 0.0224 & 0.0271 & 0.5169 & 0.5055 & 0.0111 & 0.0438 & 0.5680 & 0.5457 & 0.0945 & 0.0742 \\
\hline

    \hline

    \hline
    \end{tabular}
    }
    \label{tab:gpto3mini,moonshotv1}
\end{table*}
\begin{table*}[h]
    \centering
    \caption{Generator: DeepSeek-R1, DeepSeek-V3. "Imp.": Improvement over previous SOTA, computed as $(new - old) / (1.0 - old)$.}
    \resizebox{\linewidth}{!}{
    \begin{tabular}{l|cccc|cccc|cccc}
    \hline

    \hline

    \hline
    \multicolumn{13}{c}{\textbf{MIRAGE-DIG, DeepSeek-R1}}\\
    \hline

    \hline

    \hline
    \multirow{2}{*}{Methods}&\multicolumn{4}{c|}{Generate}&\multicolumn{4}{c|}{Polish}&\multicolumn{4}{c}{Rewrite} \\
    &  AUROC  &  Accuracy  &  MCC  &  TPR@5\%  &  AUROC  &  Accuracy  &  MCC  &  TPR@5\%  &  AUROC  &  Accuracy  &  MCC  &  TPR@5\%  \\
    \hline

    \hline
    Likelihood~\cite{likelihood} & 0.3248 & 0.5005 & 0.0224 & 0.0010 & 0.3533 & 0.5000 & 0.0000 & 0.0100 & 0.3285 & 0.5000 & 0.0000 & 0.0093 \\
    LogRank~\cite{logrank} & 0.2896 & 0.5005 & 0.0224 & 0.0010 & 0.3235 & 0.5000 & 0.0000 & 0.0089 & 0.3037 & 0.5000 & 0.0000 & 0.0062 \\
    Entropy~\cite{entropy} & 0.6454 & 0.6242 & 0.2939 & 0.0902 & 0.5593 & 0.5456 & 0.1194 & 0.1000 & 0.5772 & 0.5542 & 0.1638 & 0.1373 \\
    RoBERTa-Base~\cite{roberta} $\spadesuit$ & 0.5335 & 0.5000 & 0.0000 & 0.0000 & 0.4483 & 0.5000 & 0.0000 & 0.0000 & 0.4207 & 0.5000 & 0.0000 & 0.0021 \\
    RoBERTa-Large~\cite{roberta} $\spadesuit$ & 0.2018 & 0.5000 & 0.0000 & 0.0000 & 0.2631 & 0.5000 & 0.0000 & 0.0056 & 0.3501 & 0.5000 & 0.0000 & 0.0196 \\
\hline

    \hline

    \hline
    \end{tabular}
    }
    \label{tab:deepseekr1_deepseekv3}
\end{table*}
\begin{table*}[h]
    \centering
    \caption{Generator: Claude-3.5-Haiku, 3.7-sonnet. "Imp.": Improvement over previous SOTA, computed as $(new - old) / (1.0 - old)$.}
    \resizebox{\linewidth}{!}{
    \begin{tabular}{l|cccc|cccc|cccc}
    \hline

    \hline

    \hline
    \multicolumn{13}{c}{\textbf{MIRAGE-DIG, Claude-3.5-haiku}}\\
    \hline

    \hline

    \hline
    \multirow{2}{*}{Methods}&\multicolumn{4}{c|}{Generate}&\multicolumn{4}{c|}{Polish}&\multicolumn{4}{c}{Rewrite} \\
    &  AUROC  &  Accuracy  &  MCC  &  TPR@5\%  &  AUROC  &  Accuracy  &  MCC  &  TPR@5\%  &  AUROC  &  Accuracy  &  MCC  &  TPR@5\%  \\
    \hline

    \hline
    Likelihood~\cite{likelihood} & 0.3121 & 0.5000 & 0.0000 & 0.0000 & 0.2962 & 0.5000 & 0.0000 & 0.0000 & 0.2763 & 0.5000 & 0.0000 & 0.0011 \\
    LogRank~\cite{logrank} & 0.3153 & 0.5000 & 0.0000 & 0.0000 & 0.2761 & 0.5000 & 0.0000 & 0.0000 & 0.2592 & 0.5000 & 0.0000 & 0.0011 \\
    Entropy~\cite{entropy} & 0.7963 & 0.7320 & 0.4736 & 0.2660 & 0.7422 & 0.6918 & 0.3840 & 0.1875 & 0.7518 & 0.6762 & 0.3784 & 0.2338 \\
    RoBERTa-Base~\cite{roberta} $\spadesuit$ & 0.5036 & 0.5090 & 0.0545 & 0.0510 & 0.4503 & 0.5000 & 0.0000 & 0.0098 & 0.4400 & 0.5000 & 0.0000 & 0.0103 \\
    RoBERTa-Large~\cite{roberta} $\spadesuit$ & 0.4095 & 0.5005 & 0.0049 & 0.0370 & 0.3540 & 0.5000 & 0.0000 & 0.0110 & 0.3947 & 0.5000 & 0.0000 & 0.0194 \\
\hline

    \hline

    \hline
    \end{tabular}
    }
    \label{tab:claude3.5haiku_claude3.7sonnet}
\end{table*}
\begin{table*}[h]
    \centering
    \caption{Generator: Gemini-2.0-flash, flash-lite. "Imp.": Improvement over previous SOTA, computed as $(new - old) / (1.0 - old)$.}
    \resizebox{\linewidth}{!}{
    \begin{tabular}{l|cccc|cccc|cccc}
    \hline

    \hline

    \hline
    \multicolumn{13}{c}{\textbf{MIRAGE-DIG, Gemini-2.0-flash}}\\
    \hline

    \hline

    \hline
    \multirow{2}{*}{Methods}&\multicolumn{4}{c|}{Generate}&\multicolumn{4}{c|}{Polish}&\multicolumn{4}{c}{Rewrite} \\
    &  AUROC  &  Accuracy  &  MCC  &  TPR@5\%  &  AUROC  &  Accuracy  &  MCC  &  TPR@5\%  &  AUROC  &  Accuracy  &  MCC  &  TPR@5\%  \\
    \hline

    \hline
    Likelihood~\cite{likelihood} & 0.4471 & 0.5020 & 0.0052 & 0.0193 & 0.4069 & 0.5000 & 0.0000 & 0.0197 & 0.3967 & 0.5000 & 0.0000 & 0.0054 \\
    LogRank~\cite{logrank} & 0.4445 & 0.5000 & 0.0000 & 0.0234 & 0.3876 & 0.5000 & 0.0000 & 0.0219 & 0.3837 & 0.5000 & 0.0000 & 0.0075 \\
    Entropy~\cite{entropy} & 0.6913 & 0.6514 & 0.3071 & 0.1626 & 0.6132 & 0.5864 & 0.2002 & 0.1488 & 0.6299 & 0.5927 & 0.2471 & 0.1726 \\
    RoBERTa-Base~\cite{roberta} $\spadesuit$ & 0.4250 & 0.5010 & 0.0101 & 0.0274 & 0.4100 & 0.5000 & 0.0000 & 0.0175 & 0.4409 & 0.5000 & 0.0000 & 0.0214 \\
    RoBERTa-Large~\cite{roberta} $\spadesuit$ & 0.2978 & 0.5005 & 0.0225 & 0.0091 & 0.4322 & 0.5000 & 0.0000 & 0.0208 & 0.4876 & 0.5032 & 0.0232 & 0.0482 \\
\hline

    \hline

    \hline
    \end{tabular}
    }
    \label{tab:gemini2.0flash_flashlite}
\end{table*}
\begin{table*}[h]
    \centering
    \caption{Generator: Doubao1.5pro, Grok2. "Imp.": Improvement over previous SOTA, computed as $(new - old) / (1.0 - old)$.}
    \resizebox{\linewidth}{!}{
    \begin{tabular}{l|cccc|cccc|cccc}
    \hline

    \hline

    \hline
    \multicolumn{13}{c}{\textbf{MIRAGE-DIG, Doubao1.5pro}}\\
    \hline

    \hline

    \hline
    \multirow{2}{*}{Methods}&\multicolumn{4}{c|}{Generate}&\multicolumn{4}{c|}{Polish}&\multicolumn{4}{c}{Rewrite} \\
    &  AUROC  &  Accuracy  &  MCC  &  TPR@5\%  &  AUROC  &  Accuracy  &  MCC  &  TPR@5\%  &  AUROC  &  Accuracy  &  MCC  &  TPR@5\%  \\
    \hline

    \hline
    Likelihood~\cite{likelihood} & 0.4877 & 0.5291 & 0.0707 & 0.0040 & 0.4548 & 0.5073 & 0.0258 & 0.0093 & 0.4377 & 0.5000 & 0.0000 & 0.0105 \\
    LogRank~\cite{logrank} & 0.5231 & 0.5486 & 0.1033 & 0.0100 & 0.4496 & 0.5027 & 0.0258 & 0.0093 & 0.4395 & 0.5000 & 0.0000 & 0.0105 \\
    Entropy~\cite{entropy} & 0.6627 & 0.6249 & 0.2944 & 0.1073 & 0.5673 & 0.5650 & 0.1887 & 0.1273 & 0.5816 & 0.5634 & 0.2068 & 0.1327 \\
    RoBERTa-Base~\cite{roberta} $\spadesuit$ & 0.4880 & 0.5045 & 0.0391 & 0.0421 & 0.4709 & 0.5000 & 0.0000 & 0.0305 & 0.5464 & 0.5314 & 0.0757 & 0.0780 \\
    RoBERTa-Large~\cite{roberta} $\spadesuit$ & 0.4792 & 0.5025 & 0.0448 & 0.0371 & 0.5404 & 0.5345 & 0.0690 & 0.0584 & 0.6026 & 0.5669 & 0.1467 & 0.1013 \\
\hline

    \hline

    \hline
    \end{tabular}
    }
    \label{tab:doubao1.5pro_grok2}
\end{table*}
\begin{table*}[h]
    \centering
    \caption{Generator: Qwen2.5-7B/R1-Distill. "Imp.": Improvement over previous SOTA, computed as $(new - old) / (1.0 - old)$.}
    \resizebox{\linewidth}{!}{
    \begin{tabular}{l|cccc|cccc|cccc}
    \hline

    \hline

    \hline
    \multicolumn{13}{c}{\textbf{MIRAGE-DIG, Qwen2.5-7B-Instruct}}\\
    \hline

    \hline

    \hline
    \multirow{2}{*}{Methods}&\multicolumn{4}{c|}{Generate}&\multicolumn{4}{c|}{Polish}&\multicolumn{4}{c}{Rewrite} \\
    &  AUROC  &  Accuracy  &  MCC  &  TPR@5\%  &  AUROC  &  Accuracy  &  MCC  &  TPR@5\%  &  AUROC  &  Accuracy  &  MCC  &  TPR@5\%  \\
    \hline

    \hline
    Likelihood~\cite{likelihood} & 0.6303 & 0.6260 & 0.2634 & 0.0260 & 0.4915 & 0.5223 & 0.0460 & 0.0319 & 0.4605 & 0.5026 & 0.0063 & 0.0201 \\
    LogRank~\cite{logrank} & 0.6513 & 0.6375 & 0.2755 & 0.0312 & 0.4833 & 0.5176 & 0.0370 & 0.0277 & 0.4537 & 0.5000 & 0.0000 & 0.0180 \\
    Entropy~\cite{entropy} & 0.6164 & 0.6078 & 0.2305 & 0.0771 & 0.5493 & 0.5441 & 0.1845 & 0.1138 & 0.5729 & 0.5618 & 0.1744 & 0.0993 \\
    RoBERTa-Base~\cite{roberta} $\spadesuit$ & 0.6959 & 0.6568 & 0.3370 & 0.2948 & 0.5456 & 0.5324 & 0.0771 & 0.0596 & 0.5699 & 0.5665 & 0.1437 & 0.0813 \\
    RoBERTa-Large~\cite{roberta} $\spadesuit$ & 0.6896 & 0.6448 & 0.2898 & 0.1969 & 0.6368 & 0.6117 & 0.2237 & 0.1074 & 0.6536 & 0.6008 & 0.2046 & 0.1341 \\
\hline

    \hline

    \hline
    \end{tabular}
    }
    \label{tab:qwen2_r1distill}
\end{table*}
\begin{table*}[h]
    \centering
    \caption{Generator: LlaMa3.1-8B/R1-Distill. "Imp.": Improvement over previous SOTA, computed as $(new - old) / (1.0 - old)$.}
    \resizebox{\linewidth}{!}{
    \begin{tabular}{l|cccc|cccc|cccc}
    \hline

    \hline

    \hline
    \multicolumn{13}{c}{\textbf{MIRAGE-DIG, LlaMa3.1-8B-Instruct}}\\
    \hline

    \hline

    \hline
    \multirow{2}{*}{Methods}&\multicolumn{4}{c|}{Generate}&\multicolumn{4}{c|}{Polish}&\multicolumn{4}{c}{Rewrite} \\
    &  AUROC  &  Accuracy  &  MCC  &  TPR@5\%  &  AUROC  &  Accuracy  &  MCC  &  TPR@5\%  &  AUROC  &  Accuracy  &  MCC  &  TPR@5\%  \\
    \hline

    \hline
    Likelihood~\cite{likelihood} & 0.7698 & 0.7395 & 0.4823 & 0.1247 & 0.6213 & 0.6195 & 0.2509 & 0.0260 & 0.5904 & 0.5988 & 0.2174 & 0.0186 \\
    LogRank~\cite{logrank} & 0.7944 & 0.7588 & 0.5212 & 0.1865 & 0.6093 & 0.6008 & 0.2217 & 0.0228 & 0.5811 & 0.5837 & 0.1864 & 0.0221 \\
    Entropy~\cite{entropy} & 0.5764 & 0.5690 & 0.1623 & 0.0552 & 0.5324 & 0.5545 & 0.1345 & 0.0374 & 0.5726 & 0.5756 & 0.1799 & 0.0616 \\
    RoBERTa-Base~\cite{roberta} $\spadesuit$ & 0.8327 & 0.7682 & 0.5403 & 0.5121 & 0.5748 & 0.5496 & 0.1175 & 0.1073 & 0.6350 & 0.6099 & 0.2532 & 0.2035 \\
    RoBERTa-Large~\cite{roberta} $\spadesuit$ & 0.7945 & 0.7246 & 0.4697 & 0.4161 & 0.6461 & 0.6220 & 0.2495 & 0.0943 & 0.6999 & 0.6506 & 0.3040 & 0.1651 \\
\hline

    \hline

    \hline
    \end{tabular}
    }
    \label{tab:llama3.1_r1distill}
\end{table*}
\begin{table*}[h]
    \centering
    \caption{Generator: QwQ-Plus-32B. "Imp.": Improvement over previous SOTA, computed as $(new - old) / (1.0 - old)$.}
    \resizebox{\linewidth}{!}{
    \begin{tabular}{l|cccc|cccc|cccc}
    \hline

    \hline

    \hline
    \multicolumn{13}{c}{\textbf{MIRAGE-DIG, QwQ-Plus-32B}}\\
    \hline

    \hline

    \hline
    \multirow{2}{*}{Methods}&\multicolumn{4}{c|}{Generate}&\multicolumn{4}{c|}{Polish}&\multicolumn{4}{c}{Rewrite} \\
    &  AUROC  &  Accuracy  &  MCC  &  TPR@5\%  &  AUROC  &  Accuracy  &  MCC  &  TPR@5\%  &  AUROC  &  Accuracy  &  MCC  &  TPR@5\%  \\
    \hline

    \hline
    Likelihood~\cite{likelihood} & 0.3957 & 0.5083 & 0.0568 & 0.0010 & 0.4576 & 0.5065 & 0.0255 & 0.0156 & 0.3788 & 0.5000 & 0.0000 & 0.0068 \\
    LogRank~\cite{logrank} & 0.3715 & 0.5062 & 0.0525 & 0.0041 & 0.4298 & 0.5006 & 0.0255 & 0.0130 & 0.3553 & 0.5000 & 0.0000 & 0.0068 \\
    Entropy~\cite{entropy} & 0.6403 & 0.6165 & 0.2723 & 0.0714 & 0.5070 & 0.5234 & 0.0806 & 0.0519 & 0.5786 & 0.5609 & 0.1837 & 0.1229 \\
    RoBERTa-Base~\cite{roberta} $\spadesuit$ & 0.5096 & 0.5000 & 0.0000 & 0.0021 & 0.4165 & 0.5000 & 0.0000 & 0.0052 & 0.4212 & 0.5000 & 0.0000 & 0.0080 \\
    RoBERTa-Large~\cite{roberta} $\spadesuit$ & 0.2154 & 0.5000 & 0.0000 & 0.0010 & 0.3700 & 0.5000 & 0.0000 & 0.0091 & 0.4054 & 0.5000 & 0.0000 & 0.0307 \\
\hline

    \hline

    \hline
    \end{tabular}
    }
    \label{tab:qwq-plus}
\end{table*}
\end{document}